%% file: main.tex
\documentclass[sigconf, nonacm, natbib=false]{acmart}

\usepackage{amsmath}
\usepackage{caption}
\usepackage{subcaption}
\usepackage{cleveref}
\usepackage{dsfont}
\usepackage{siunitx}
\usepackage{booktabs}
\usepackage[mode=buildnew]{standalone}
\usepackage{lipsum}
\usepackage{xcolor}
\usepackage{siunitx}
\usepackage{bm}
\usepackage{tikz}
\usepackage{pgfplots}
\usepackage{algorithm}
\usepackage{algorithmicx}
\usepackage[noend]{algpseudocode}
\usepackage{vector} 

\input{defs}
\input{figures/preamble}

\AtBeginDocument{%
  }

\setcopyright{acmcopyright}
\copyrightyear{2018}
\acmYear{2018}
\acmDOI{XXXXXXX.XXXXXXX}

\RequirePackage[
  datamodel=acmdatamodel,
  style=acmnumeric,
  ]{biblatex}

\begin{document}

\title{Entropy-regularized Point-based Value Iteration}

\author{Harrison Delecki}
\email{hdelecki@stanford.edu}
\affiliation{%
  \institution{Stanford University}
  \streetaddress{496 Lomita Mall}
  \city{Stanford}
  \state{California}
  \country{USA}
  \postcode{94305}
}

\author{Marcell Vazquez-Chanlatte}
\email{marcell.chanlatte@nissan-usa.com}
\affiliation{%
  \institution{Nissan Advanced Technology Center Silicon Valley}
  \streetaddress{ 3400 Central Expy}
  \city{Santa Clara}
  \state{California}
  \country{USA}
  \postcode{95051}
}

\author{Esen Yel}
\email{esenyel@stanford.edu}
\affiliation{%
  \institution{Stanford University}
  \streetaddress{496 Lomita Mall}
  \city{Stanford}
  \state{California}
  \country{USA}
  \postcode{94305}
}

\author{Kyle Wray}
\email{kylewray@stanford.edu}
\affiliation{%
  \institution{Stanford University}
  \streetaddress{496 Lomita Mall}
  \city{Stanford}
  \state{California}
  \country{USA}
  \postcode{94305}
}


\author{Tomer Arnon}
\email{tomer.arnon@nissan-usa.com}
\affiliation{%
  \institution{Nissan Advanced Technology Center Silicon Valley}
  \streetaddress{ 3400 Central Expy}
  \city{Santa Clara}
  \state{California}
  \country{USA}
  \postcode{95051}
}

\author{Stefan Witwicki}
\email{stefan.witwicki@nissan-usa.com}
\affiliation{%
  \institution{Nissan Advanced Technology Center Silicon Valley}
  \streetaddress{ 3400 Central Expy}
  \city{Santa Clara}
  \state{California}
  \country{USA}
  \postcode{95051}
}

\author{Mykel J. Kochenderfer}
\email{mykel@stanford.edu}
\affiliation{%
  \institution{Stanford University}
  \streetaddress{496 Lomita Mall}
  \city{Stanford}
  \state{California}
  \country{USA}
  \postcode{94305}
}




\renewcommand{\shortauthors}{Delecki et al.}

\begin{abstract}
Model-based planners for partially observable problems must accommodate both model uncertainty during planning and goal uncertainty during objective inference. However, model-based planners may be brittle under these types of uncertainty because they rely on an exact model and tend to commit to a single optimal behavior. Inspired by results in the model-free setting, we propose an entropy-regularized model-based planner for partially observable problems. Entropy regularization promotes policy robustness for planning and objective inference by encouraging policies to be no more committed to a single action than necessary. We evaluate the robustness and objective inference performance of entropy-regularized policies in three problem domains. Our results show that entropy-regularized policies outperform non-entropy-regularized baselines in terms of higher expected returns under modeling errors and higher accuracy during objective inference.
\end{abstract}




\maketitle

\section{Introduction}

Recent applications of partially observable Markov decision processes (POMDPs) have shown success in a range of applications such as autonomous vehicles \cite{wray2021pomdps}, robotic planning \cite{lauri2022partially}, aircraft collision avoidance \cite{kochenderfer2012next}, and objective inference \cite{ramirez2011goal}. Model-based planners should be robust to the uncertainty that arises in these different domains. However, modeling errors may cause model-based methods to struggle because they rely on exact knowledge of dynamics and observation models. Additionally, planners are often part of larger decision-making systems that are not included in the model \cite{wray2017modia}. When inferring another agent's objective, there are often multiple strategies the agent could use to achieve a certain goal. Model-based policies for objective inference are likely to underperform because they over-commit to optimal behaviors \cite{ziebart2008maximum}. In this work, we propose an entropy-regularizing objective and POMDP planning algorithm to address these issues.



Entropy-regularized objectives are frequently used in model-based planning in a fully observable context \cite{grill2019planning},  model-free reinforcement learning \cite{haarnoja2018soft}, and inverse reinforcement learning (IRL) \cite{ziebart2008maximum}. In model-free and fully observable planning settings, entropy regularization promotes policy robustness by encouraging the policy to be no more committed to a single optimal action than necessary. In the IRL setting, entropy regularization helps enable inference over multiple, possibly suboptimal strategies for meeting an objective. Additionally, entropy-regularization yields smoother value functions, which can facilitate optimization~\cite{grill2019planning}. While previous applications of entropy-regularization have found success in fully observable domains, there has been relatively less study of entropy-regularization in the partially observed setting. 



Some previous works on entropy-regularized planning introduce a reward bonus based on the information entropy of the agent's belief \cite{curtis2023taskerb} and maximizing the predictability of the agent's behavior~\cite{molloy2023entropy}. These instances of entropy regularization aim to promote observability in the agent's belief and behavior, respectively. In contrast, our goal is to encourage the planner to consider multiple potential behaviors rather than overcommitting to a single one. A previous approach similar to ours maximizes policy entropy subject to reward-based constraints by synthesizing a finite state controller \cite{savas2022entropy}. Rather than introducing constraints, we introduce a soft regularization term to promote policy entropy. Additionally, we consider belief-based alpha vector policies.


Since the value function of POMDPs is piecewise-linear and convex, point-based value iteration (PBVI) \cite{pineau2003point} and many other POMDP planning algorithms \cite{spaan2005perseus, kurniawati2008sarsop} use alpha vectors to represent approximate value functions and policies.  Alpha vector policies are desirable for their auditability and low computational requirements in online applications. Previous work has also investigated entropy-regularized alpha vector policies in the context of QMDP, which uses a fully observable approximation of the full POMDP \cite{park2022anderson}. This approximation may lead to poor performance, especially when the POMDP involves information-gathering actions. We propose an entropy-regularized planning algorithm for the POMDP without the fully observable approximation of QMDP.


In this work, we propose an entropy-regularized objective for model-based planning and an offline point-based solver, Entropy-regularized Point-based Value Iteration (ERPBVI).  We evaluate the robustness and objective inference performance of \erpbvi policies in toy problems and an automotive-inspired example. Our experiments show that our entropy-regularized policies are more robust to modeling errors and perform more accurate objective inference than non-entropy-regularized baselines.

\section{Background}

This section provides the necessary background on POMDPs, belief-state MDPs, and Point-based Value Iteration.

\subsection{POMDPs}
A POMDP is a formulation for sequential decision-making problems where the underlying state is unobservable. POMDPs are represented by the tuple $\langle\mathcal{S}, \mathcal{A}, \mathcal{O}, T, O, R, \gamma \rangle$ where $\mathcal{S}$, $\mathcal{A}$, $\mathcal{O}$ denote the state, action, and observation spaces, respectively. Given a current state $s\in\mathcal{S}$ and agent action $a\in\mathcal{A}$, the agent transitions to a new state $s'$ according to the transition model $s'\sim T\left(\cdot \mid s, a \right)$. The agent cannot observe the true state of the system, but receives an observation $o \sim O\left(\cdot \mid s', a\right)$. The agent uses this observation to update its current belief $b$ over possible states to obtain a posterior distribution $b'(s)$:
\begin{equation*}
    b'(s') \propto O(o \mid s,a) \sum_{s\in\mathcal{S}} T(s' \mid s, a)b(s)
\end{equation*}
For compactness, we denote the belief update process using the operation $b' = \update(b, a, o)$.

A stochastic policy $\pi(a \mid b)$ is a distribution over actions given a current belief $b$. The agent receives a reward specified by $R(s, a)$ for taking action $a$ in state $s$.

\subsection{Belief-state Markov Decision Process}
Any POMDP can be converted into an MDP that uses beliefs as states, also known as a belief-state MDP \cite{kochenderfer2022algorithms}. In a belief-state MDP, the state space is continuous because it represents the set of all possible beliefs. The action space remains the same as the POMDP. The reward function becomes an expected value of the state-based reward. For a discrete state and action space, the belief-state MDP reward is:
\begin{equation*}
    R(b, a) = \sum_{s\in b} b(s) R(s, a)
\end{equation*}
The belief-state transition function $T(b' \mid b, a)$ is given by:
\begin{equation*}
    s \sim b(\cdot) \quad s' \sim T(\cdot \mid s, a) \quad o \sim O(\cdot \mid s', a) \quad b' = \update(b, a, o)
\end{equation*}

\subsection{Point-based Value Iteration}
Point-based Value Iteration (PBVI) is an offline approximate solution method for POMDPs~\cite{pineau2003point}. PBVI and other offline solvers represent an approximate value function using a collection of vectors known as alpha vectors. The value function for POMDPs is piecewise-linear and convex~\cite{smallwood1973optimal}. PBVI approximates the value function by a set of $m$ vectors $\Gamma=\{\vec{\alpha}_1,\ldots, \vec{\alpha}_m\}$. Each alpha vector $\vec{\alpha}_i$ defines a hyperplane in belief space. The approximate value function is defined by:
\begin{equation*}
    U^{\Gamma}(\vec{b}) = \max_{\vect{\alpha} \in \Gamma} \vec{\alpha}^{\top} \vec{b}
\end{equation*}

PBVI computes a set $\Gamma$ of $m$ different alpha vectors associated with $m$ different belief points $B=\{\vec{b}_1,\ldots, \vec{b}_m\}$. The vectors in $\Gamma$ are initialized to a lower bound on the optimal value function. The algorithm proceeds by alternating between a backup step and an expansion step.

The backup step takes a belief and a set of alpha vectors $\Gamma$ and constructs a new alpha vector to improve the approximate value function. The backup iterates over every possible action $a$ and observation $o$, and extracts the maximal alpha vector from the next belief state:
\begin{equation*}
    \vec{\alpha}_{a,o} = \argmax_{\vec{\alpha} \in \Gamma} \vec{\alpha}^{\top} \update(\vec{b}, a, o)
\end{equation*}
The algorithm computes a new alpha vector for each available action based on these $\alpha_{a,o}$ vectors, given by:
\begin{equation*}
    \vec{\alpha}_a(s) = R(s, a) + \gamma \sum_{s'\in \mathcal{S}} \sum_{o \in \mathcal{O}} O(o\mid s', a) T(s' \mid s, a) \vec{\alpha}_{a,o}(s')
\end{equation*}
The backed-up alpha vector at belief $\vec{b}$ is:
\begin{equation*}
    \vec{\alpha} = \argmax_{\vec{\alpha}_a} \vec{\alpha}_a^{\top} \vec{b}
\end{equation*}
The expansion step grows the set of belief points $B$ to cover more points in belief space. The algorithm computes new candidate belief points by computing every action and observation from each point in $B$. For each belief in $B$, the expanded belief point that is the furthest away from all other points in $B$ is added to $B$.

 The lower bound on the value function represented by $\Gamma$ increases with repeated application of the backup until convergence. The converged value function will not be optimal because $B$ typically does not include all beliefs.

\section{Entropy-regularized PBVI}
\begin{figure}[!tb]
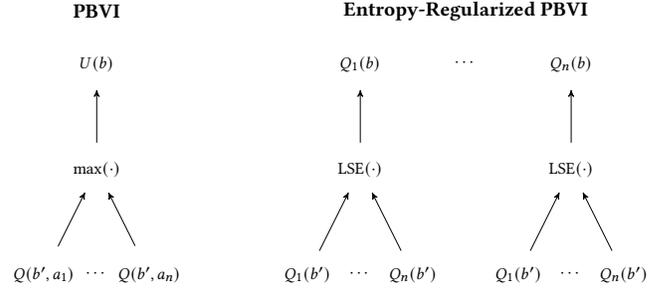

    \centering
    \includestandalone[width=\linewidth]{figures/backup}
    \caption{Illustration of the PBVI backup (left) and entropy-regularized PBVI backup (right). The PBVI updates an estimate of the value function by taking the maximum over Q-values at next belief points. The entropy-regularized variant explicitly models Q-functions for all $n$ actions, and the maximum over Q-values is replaced by $\operatorname{LogSumExp}$ (LSE).}
    \label{fig:backup}
\end{figure}
Now we introduce an entropy-regularized version of PBVI, which we call \erpbvi. First, introduce an entropy-regularized objective for planning in a belief-state MDP. Then, we describe details of the \erpbvi algorithm backup step and policy extraction.

\subsection{Entropy-regularized Planning Objective}
Here we introduce an entropy-regularized objective in the belief-MDP, inspired by previous work on entropy regularization in the fully-observable setting \cite{geistTheoryRegularizedMarkov2019a}. The entropy of a stochastic policy $\pi$ at belief $\vec{b}$ is:
\begin{equation}
\label{eq:entropy}
    H(\pi, \vec{b}) = -\sum_{a \in \mathcal{A}} \pi(a \mid \vec{b}) \log \pi(a \mid \vec{b})
\end{equation}
We define the entropy-regularized value as
\begin{equation}
    \label{eq:er-value}
    \begin{split}
    U(\vec{b}) = \max_{\pi(\cdot \mid \vec{b}) \in \mathcal{P}_{\mathcal{A}}}  &\E_{a \sim \pi(\cdot \mid \vec{b}), \vec{b}' \sim T(\cdot \mid \vec{b}, a)} \left[ R(\vec{b}, a) + \gamma  U(\vec{b}') \right]\\
    & + \lambda H(\pi, \vec{b})
    \end{split}
\end{equation}
where $\mathcal{P}_{\mathcal{A}}$ is the set of all probability distributions over $\mathcal{A}$, and $\lambda \geq 0$  and controls the degree of regularization. Since the negative entropy is the convex conjugate of the log-sum-exp function \cite{geistTheoryRegularizedMarkov2019a}, \cref{eq:er-value} is equivalent to:
\begin{equation}
    \label{eq:er-conj}
    U(\vec{b}) = \lambda \log \left( \sum_{a \in \mathcal{A}} \exp \left( \frac{1}{\lambda} \E \left[ R(\vec{b}, a) + \gamma  U(\vec{b}') \right] \right)  \right)
\end{equation}
Since the term in the expectation is equivalent to the Q-value for belief $\vec{b}$ and action $a$, we can write the entropy regularized value function as the log-sum-exp of Q-values weighted by the regularizing strength:
\begin{equation}
    \label{eq:er-qvalue}
    U(\vec{b}) = \lambda \log \left( \sum_{a \in \mathcal{A}} \exp \left( \frac{1}{\lambda} Q(\vec{b}, a) \right)  \right)
\end{equation}
For the remainder of this paper, we will use the operator $\operatorname{LogSumExp}$ to represent the log-sum-exp similar to \cref{eq:er-qvalue}.

\subsection{\erpbvi Backup}
Rather than computing an approximation of the value function at each belief point, we maintain an estimate of the Q-function for each action. This difference between the PBVI and \erpbvi backup is illustrated in \cref{fig:backup}. The Q-function for action $a$ is denoted by the collection of alpha-vectors $\Gamma_a$, and the set of all Q-functions is $\Xi = \{\Gamma_1, \ldots, \Gamma_{|\mathcal{A}|} \}$. The set $\Gamma_a$ defines an approximate Q-function for action $a$:
\begin{equation}
    \label{eq:vec-max-q}
    Q_a(\vec{b}) = \max_{\vec{\alpha} \in \Gamma_a} \vec{\alpha}^{\top} \vec{b}
\end{equation}
The value function is then identical to \cref{eq:er-qvalue}.

The algorithm maintains a lower bound on the optimal value function, $U^{\Gamma}(\vec{b}) \leq U^{*}(\vec{b})$. We initialize each approximate Q-function $\Gamma_a$ to a lower bound and perform a backup to update the alpha vectors in each $\Gamma_a$ and point in $B$. The backup takes in a belief point in $B$ and the set $\Xi$ of Q-functions $\Gamma_a$, and constructs a new alpha vector for each Q-function. For each action $a$, the algorithm iterates through observations $o$ and extracts the alpha vector from each Q-function $\Gamma_a$ that is maximal at the resulting belief state. The set of dominating alpha vectors is $\Gamma_{a, o}$, where the $i$th element is:
\begin{equation}
    \label{eq:gamma-argmax}
    \Gamma_{a, o}^{i} = \argmax_{\vec{\alpha} \in \Gamma_i} \vec{\alpha}^{\top} \operatorname{Update}(\vec{b}, a, o)
\end{equation}
For convenience, we form the $|\mathcal{A}|$ alpha vectors of $\Gamma_{a,o}$ into the columns of a matrix $\mat{A}_{a,o}$. From \cref{eq:er-qvalue}, we can write the approximate entropy-regularized utility at the next belief-state as:
\begin{equation}
    \label{eq:bprime-lse-util}
    U\left(\vec{b}' \right) = \lambda \operatorname{LogSumExp}\left(\frac{1}{\lambda} \mat{A}_{a,o}^{\top} \update(\vec{b}, a, o)\right)
\end{equation}
Next, we compute an alpha vector representing a linear approximation of the utility at the next belief state. Since the $\operatorname{LogSumExp}$ function is differentiable, we can compute the gradient of \cref{eq:bprime-lse-util} which gives an alpha vector $\vec{\alpha}_{a,o}$ corresponding to the entropy regularized utility at the next belief:
\begin{equation}
    \label{eq:alpha-ao-soft-update}
    \vec{\alpha}_{a,o} = \mat{A}_{a,o} \operatorname{SoftMax} \left( \frac{1}{\lambda}\mat{A}_{a,o}^{\top} \operatorname{Update}(\vec{b}, a, o) \right)
\end{equation}
We construct one new alpha vector for each Q-function using all $\vec{\alpha}_{a,o}$ and a one-step lookahead:
\begin{equation}
    \label{eq:alpha-lookahead}
    \vec{\alpha}_a(s) = R(s, a) + \gamma \sum_{s',o} O(o \mid a, s') T(s' \mid s, a) \vec{\alpha}_{a,o}(s')
\end{equation}
The new alpha vectors are combined with the existing $\Gamma_a$. Each approximate Q-function is pruned after every backup to remove dominated alpha vectors. Alpha vector pruning is performed as described by \cite{wray2021pomdps}\citeauthor{wray2021pomdps}. The remainder of the algorithm is the same as PBVI~\cite{pineau2003point}. Pseudocode for \erpbvi is available in \cref{chapter:algorithms}.

\subsection{Policy Extraction}
We extract a stochastic policy based on the entropy-regularized objective inspired by previous work in the fully-observable setting \cite{geistTheoryRegularizedMarkov2019a, grill2019planning}. The gradient of the objective in \cref{eq:er-qvalue} defines a softmax distribution over actions where the probability of action $a_i$:
\begin{equation}
    \label{eq:er-policy}
    \pi(a_i \mid \vec{b}) = \frac{\exp(Q_i(\vect{b})/\lambda)}{\sum_{j=1}^{|\mathcal{A}|} \exp(Q_j(\vect{b})/\lambda)}
\end{equation}
This is the form of the \erpbvi policy we use in experiments.

\section{Experiments}

This section discusses experiments designed to evaluate entropy-regularized PBVI policies in robustness and objective inference performance. We describe POMDP problems used in all experiments and details of each experiment.

\subsection{Robustness Experiments}
Robustness experiments evaluate the sensitivity of \erpbvi policies to modeling errors. First, we train a policy on a POMDP problem with known parameters. We evaluate the robustness of policies to environment changes by estimating the expected sum of discounted rewards $\E\left[R \right]$ on the same POMDP problem. As a baseline, we consider non-entropy-regularized PBVI policies. Next, we describe POMDP problems used to evaluate robustness.

\subsubsection{Tiger}

We evaluate the robustness of a trained policy to observation noise in the Tiger POMDP by changing the probability of listening correctly for the tiger. The training POMDP uses the observation model as presented in \cite{cassandra1994acting}. If the tiger is behind the left door, then the agent will hear the tiger correctly on the left side with probability $p_{correct}=0.85$. We evaluate the trained policy on problems with $p_{correct}=0.6$, $p_{correct}=0.7$ and $p_{correct}=0.9$. If the agent opens the door and finds the tiger, it gets a reward of \num{-100}. If the agent opens the other door, it escapes and receives a reward of \num{10}. The agent receives a reward of \num{-1} for listening. 
We compute the expected sum of discounted returns using $100$ rollouts with horizon $100$. For \erpbvi, we train $30$ policies with temperature $\lambda$ varying between $10^{-2}$ and $10^{2}$ evenly spaced on a log scale.

\subsubsection{GridWorld}
\begin{figure}[!tbp]
  \begin{subfigure}[b]{0.35\linewidth}
    \includegraphics[width=\linewidth]{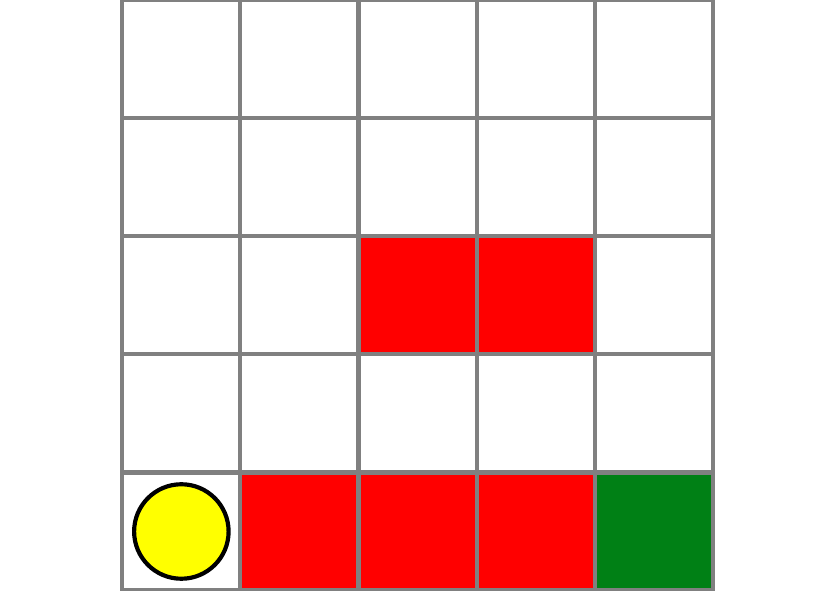}
  \end{subfigure}
  \hfill
  \begin{subfigure}[b]{0.3\linewidth}
    \includestandalone[width=\linewidth]{figures/pbvistatedist}
  \end{subfigure}
  \hfill
  \begin{subfigure}[b]{0.3\linewidth}
    \includestandalone[width=\linewidth]{figures/erstatedist}
  \end{subfigure}
  \caption{Illustration of the GridWorld POMDP (left). The goal agent tries to reach the lower-right corner while avoiding failure states. The expected visitation rate for the PBVI policy (middle) shows that the agent follows a single path to the goal through the narrow passage. The ERPBVI policy (right) follows multiple paths to the goal.}
  \label{fig:gridworld-overview}
\end{figure}

We also consider policy robustness under transition model uncertainty in a GridWorld problem inspired by problems in \cite{littman1995learning}. States correspond to each location in the grid. The agent can try to move in the four cardinal directions. If the agent hits a wall, the agent stays in the current location. The GridWorld is `slippery', meaning that there is some probability $p_{slip}$ that the agent will remain in the current location.

When the agent takes an action and the next cell is free, it will slip with probability $p_{slip}$ and remain at the current location. Otherwise, the agent moves to the next location. The grid is configured as shown in \cref{fig:gridworld-overview}. The green cell in the lower right corner represents a goal state, where the agent receives a reward of $+1$ and deterministically transitions to a terminal state. Red squares represent failure states, where the agent receives a $-1$ reward before deterministically transitioning to the terminal state. The agent can either take a short, risky path or a longer safer path to the goal. There are three observations, corresponding to the three types of grid states (normal, goal, and failure). The agent always starts in the bottom left corner. 

We train policies with $p_{slip}=0$, and evaluate robustness on $p_{slip}=0.1$, $p_{slip}=0.3$, and $p_{slip}=0.5$. We evaluate the robustness of \erpbvi policies with $30$ temperatures between $10^{-4}$ and $10^{2}$ evenly spaced on a log scale.

\subsection{Objective Inference Experiments}
The objective inference experiments are inspired by previous work in IRL and goal recognition \cite{ziebart2008maximum, ramirez2011goal}. Suppose we observe an agent acting in a POMDP with a set $\mathcal{G}$ of possible goals $G$. For example, there may be multiple goal locations in a simple navigation domain. We assume that we observe the agent's action-observation sequence $\tau=\left[a_1, o_1, \ldots,a_t, o_t \right]$ up to time step $t$. We also assume that the agent's initial belief $\vec{b}_0$ is known, so that we can compute the agent's belief state at any step.

Our goal is to compute the posterior probabilities $P(G \mid \tau)$. Assuming a uniform prior over goals, the inference query $P(G \mid \tau)$ can be expanded:
\begin{equation}
    P(G \mid \tau) \propto P(\tau \mid G)
\end{equation}
A key assumption to approximate $P(\tau \mid G)$ is that if the agent is pursuing goal $G$, then the agent will choose action $a$ in belief-state $\vec{b}$ with probability
\begin{equation}
    P(a \mid \vec{b}, G) = \pi_G(a \mid \vec{b})
\end{equation}
where $\pi_G$ is a stochastic policy trained to maximize the objective for goal $G$. We can approximate the probability of the action-observation sequence $\tau$ given goal $G$ as
\begin{equation}
    P(\tau \mid G) \approx \prod_{i=1}^t \pi_G \left(a_i \mid \vec{b}_i\right)
\end{equation}
where $\vec{b}_i$ is the belief-state computed from $\tau$ for time step $i$. After computing the posterior $P(G \mid \tau)$ for each goal, we take the maximum likelihood estimate to predict $G$.

To evaluate the objective inference performance, we compare the true-positive rate, false-positive rate, and accuracy of using \erpbvi and PBVI policies. We train \erpbvi policies for each goal $G$, and use the stochastic policy from \cref{eq:er-policy} to compute action probabilities. As a baseline, we build a stochastic policy from a softmax distribution over non-entropy regularized PBVI Q-values 
\begin{equation}
    \label{eq:pbvi-policy}
    \pi^{PBVI}(a \mid \vec{b}) \propto \exp(Q(\vec{b}, a)/ \lambda)
\end{equation}
where $Q(\vec{b}, a)$ is the action-value using the PBVI alpha vectors and $\lambda$ is a temperature parameter. Next, we describe the GridWorld and Crosswalk POMDPs used in objective inference experiments.

\begin{figure}[!tb]
  \begin{subfigure}[b]{0.4\linewidth}
    \includegraphics[width=\linewidth]{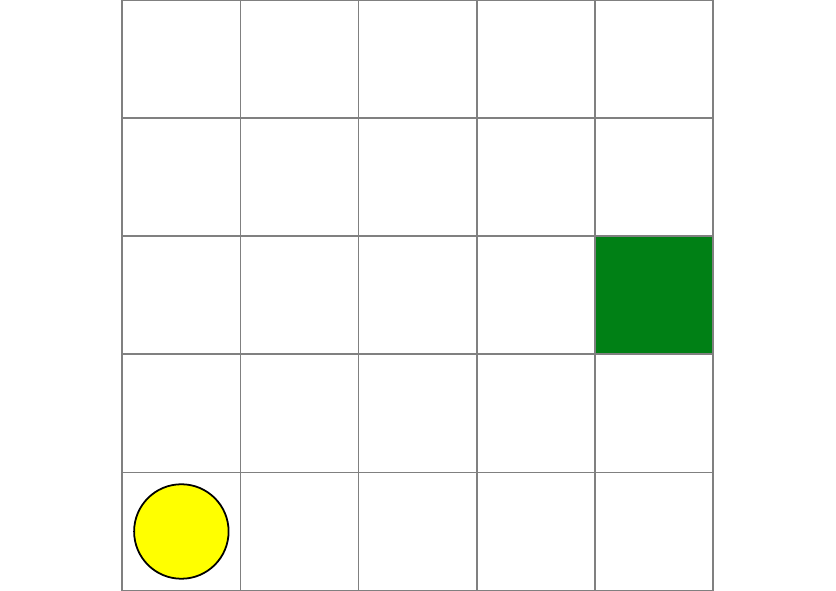}
    \label{fig:grid-obj-1}
  \end{subfigure}
  \hfill
  \begin{subfigure}[b]{0.4\linewidth}
    \includegraphics[width=\linewidth]{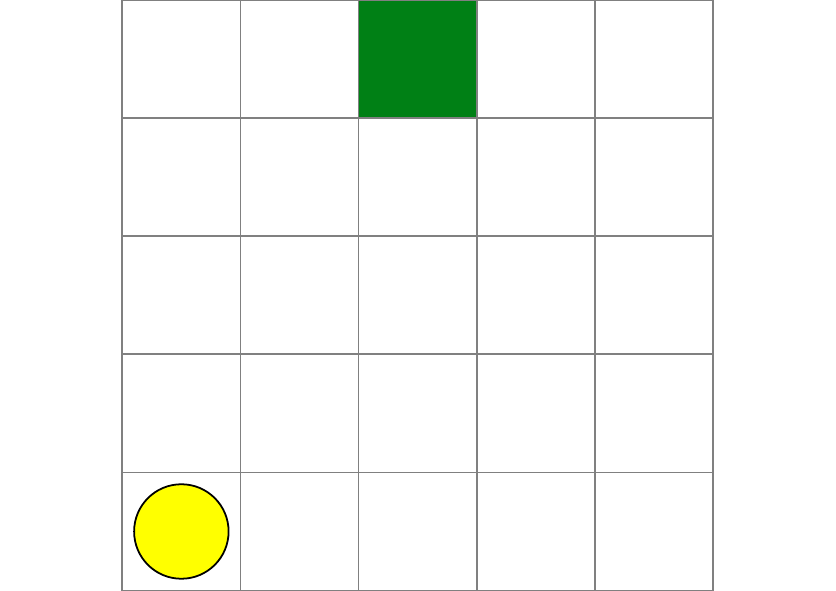}
    \label{fig:grid-obj-2}
  \end{subfigure}
  \caption{Illustration of the two GridWorld planning objectives used in the objective inference experiments. The states, actions, and observations are identical in each POMDP.}
  \label{fig:gridworld-inference-problems}
\end{figure}

\subsubsection{GridWorld} We perform objective inference in a GridWorld POMDP identical to the one presented in robustness experiments but without the failure cells. The two objectives, visualized in \cref{fig:gridworld-inference-problems}, each have a single goal location against one of the far walls from the agent's initial position. We generate $\num{1000}$ action-observation trajectories using the deterministic PBVI policies that select random actions with probability $0.5$. Each method observes the first $\num{5}$ time steps of each trajectory. We compare each method's true-positive rate and false-positive rate for $\num{10}$ temperatures between $10^{-2}$ and $10^2$ spaced linear on a log scale.

\subsubsection{Crosswalk}

We model a pedestrian crossing the street at a crosswalk with a single oncoming car. The problem's state is a tuple $\langle p_{ped}, p_{car}, v_{car} \rangle$ where $p_{ped}$ is the position of the pedestrian along the crosswalk, $p_{car}$ is the position of the vehicle in the street perpendicular to the crosswalk, and $v_{car}$ is the vehicle's velocity.  The pedestrian's actions include moving forward, moving backward, or remaining stationary. The crosswalk and adjoining street are discretized into 10 positions.  The oncoming vehicle can be moving forward zero, one or two units per time step. The vehicle's dynamics depend solely on whether the pedestrian is in the street. When the pedestrian is on the side of the road, the oncoming vehicle brakes with probability $p_{b}=0.5$. If the pedestrian is in the street, the vehicle brakes with a larger probability $p_{c}=0.9$. The street begins at position $3$ along the crosswalk. If the pedestrian is in the street $p_{ped}>3$ and $p_{car}\in[7, 9]$, there is a collision. The agent receives noisy observations of the vehicle's position.

We use two variants of this problem for objective inference. In the first version, the pedestrian incurs a large penalty for being struck by the vehicle. This penalty prioritizes safe street-crossing behaviors, emphasizing avoidance of the oncoming car. In contrast, the second version features no penalty when the pedestrian is struck, representing a scenario where the pedestrian disregards the oncoming traffic. We use the same data generation parameters as in the GridWorld experiment. Each method observes the first $\num{10}$ time steps of each trajectory. We compare each method's true-positive rate and false-positive rate for nine temperatures between $10^{-4}$ and $10^{4}$ linearly spaced on a log scale.

\begin{figure}[!tb]
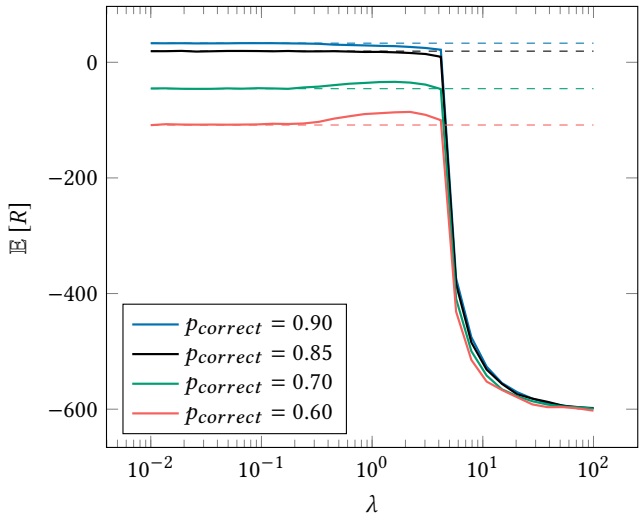

    \centering
    \includestandalone[width=\linewidth]{figures/tiger_robustness_single}
    \caption{Robustness results on Tiger POMDP using \erpbvi (solid line) and PBVI (dashed line) for various values of $p_{correct}$. The nominal $p_{correct}=0.85$ is shown in black. When the model is worse than expected ($p_{correct} \leq 0.85$), \erpbvi achieves higher expected return $\mathbb{E}[R]$ than $PBVI$ for $\lambda \approx 1$.}
    \label{fig:tiger-robustness}
\end{figure}


\section{Results}
This section presents results of the robustness and objective inference experiments using \erpbvi. 

\subsection{Robustness}
First, we present results of the robustness experiments on the Tiger and Gridworld POMDPs. The maximum improvement in expected return using \erpbvi compared to PBVI is shown in \cref{tab:robustness-summary}.

\subsubsection{Tiger}

Robustness results for the Tiger POMDP are shown in \cref{fig:tiger-robustness}. The solid line indicates \erpbvi performance, and the dashed line indicates the regular PBVI performance. The black lines show performance on the training problem. \erpbvi can only perform as well as PBVI when the evaluation problem is identical to the training problem. As the policy temperature is increased, the \erpbvi policy converges to the uniform random policy, leading to much worse performance.

When the true $p_{correct}$ is smaller than during training, there exists a range of temperatures where \erpbvi outperforms PBVI. However, when the observations are less noisy than training,  the entropy-regularized policies cannot perform any better. This is because entropy regularization leads to policies that place less trust in the observation model.

\subsubsection{GridWorld}
\begin{figure}[!tb]
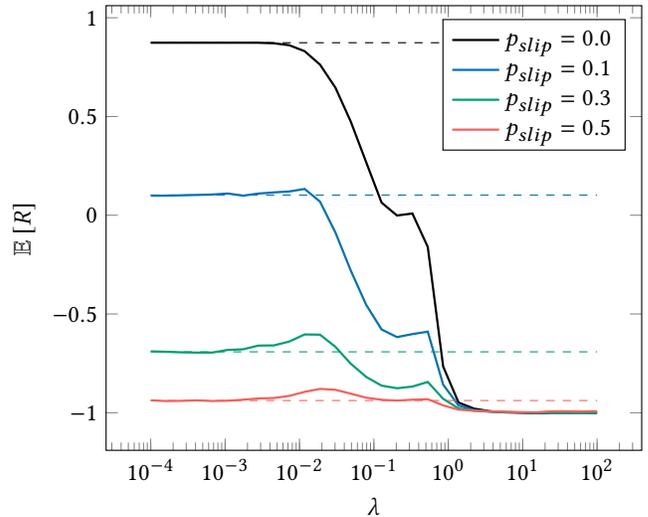

    \centering
    \includestandalone[width=\linewidth]{figures/grid_pareto}
    \caption{Expected return for PBVI and \erpbvi policies in GridWorld with varying $p_{slip}$. All policies are trained with $p_{slip}=0$. Solid lines indicate \erpbvi performance, while dashed lines indicate PBVI. \erpbvi is less sensitive to changes in the POMDP's transition model when $\lambda \approx 0.02$.}
    \label{fig:grid-robustness}
\end{figure}

\begin{table}[!tb]
  \caption{Maximum increase in mean return $\Delta$ for \erpbvi versus PBVI on Tiger and GridWorld.}
  \label{tab:robustness-summary}
    \centering
    \begin{tabular}{@{}c S S S S S S S@{}}
      \toprule
       &  \multicolumn{3}{c}{$\text{Tiger}$}  & &  \multicolumn{3}{c}{$\text{GridWorld}$} \\
       & \multicolumn{3}{c}{$p_{correct}$}  & &  \multicolumn{3}{c}{$p_{slip}$} \\
      & \num{0.9} & \num{0.7} & \num{0.6} & & \num{0.1} & \num{0.3} & \num{0.5}\\
      \midrule
      $\Delta$ & \num{0.0} & \num{11.81} & \num{22.62}  & & 0.03 & 0.09 & 0.06  \\
      \bottomrule
    \end{tabular}
\end{table}

Robustness results for the GridWorld problem are shown in \cref{fig:grid-robustness}. As before, the solid line denotes \erpbvi and the dashed line denotes PBVI. For certain temperature ranges, the entropy-regularized policy achieves a greater expected discounted return than the PBVI policy. The difference in behavior between PBVI and \erpbvi is illustrated in \cref{fig:gridworld-overview}. The PBVI policy will always try to go through the narrow corridor towards the goal, since it is trained in a deterministic environment. In contrast, the stochastic entropy-regularized policy occasionally takes both the short, risky path as well as the safer path around the failure states.

\subsection{Objective Inference}
Here we present the results of the inference experiments on the GridWorld and Crosswalk POMDPs.

\subsubsection{GridWorld} 
\begin{figure}[!tb]
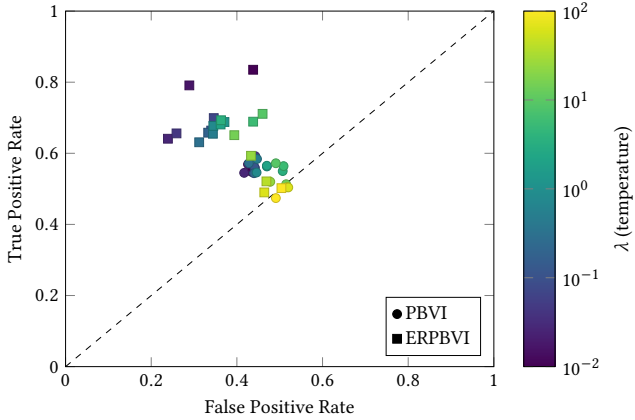

    \centering
    \includestandalone[width=\linewidth]{figures/gridworld_inference_tpr_fpr}
    \caption{True positive rate versus false positive rate for \erpbvi and PBVI policies on the GridWorld POMDP. For low temperatures, \erpbvi policies perform better (higher TPR and lower FPR) than the PBVI policies. When the temperature is large, the inference performance for both methods decays to random guessing.}
    \label{fig:grid-inference-tpr-fpr}
\end{figure}

Results for objective inference in the GridWorld are shown in \cref{fig:grid-inference-tpr-fpr}. The plot shows the true-positive rate versus false-positive rate using \erpbvi (squares) and PBVI baseline (circles) policies with a variety of temperatures. The dashed line indicates the theoretical performance of a random classifier. Assuming equal preference between true positives and false positives, the ideal performance would be in the upper-left corner. At low temperatures, \erpbvi achieves better performance than the PBVI baseline with higher true-positive rates and lower false-positive rates. Inference using \erpbvi policies is less sensitive to random or sub-optimal action sequences because the entropy-regularization encourages the policy to be less committed to a single optimal action. When the temperature increases, the inference performance of both \erpbvi and PBVI reverts to random guessing, since the high-temperature softmax induces a uniform distribution over actions.

\subsubsection{Crosswalk}
\begin{figure}[tb]
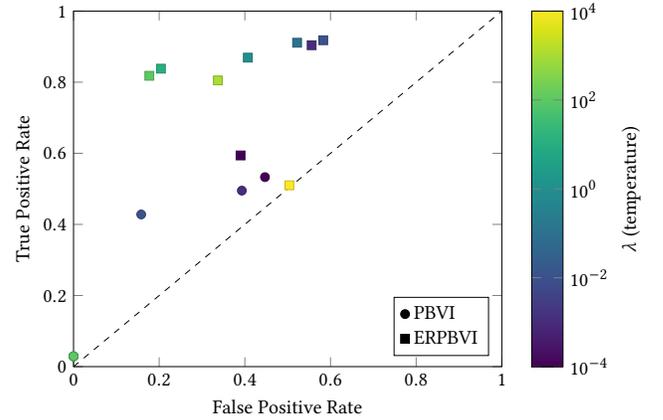

    \centering
    \includestandalone[width=\linewidth]{figures/crosswalk_tpr}
    \caption{True positive rate versus false positive rate for \erpbvi and PBVI policies on the Crosswalk POMDP. The positive classification corresponds to the safe behavior of avoiding collision. \erpbvi policies significantly outperform PBVI policies with higher true positive rate and lower false positive rate. Six of the PBVI points overlap at the lower left.}
    \label{fig:crosswalk-inference-tpr-fpr}
\end{figure}

True positive rate versus false positive rate for the Crosswalk problem is shown in \cref{fig:crosswalk-inference-tpr-fpr}. \erpbvi significantly outperforms PBVI for some temperature ranges with higher true positive rates and lower false positive rates. The difference in performance is more dramatic in this problem because PBVI overcommits to a single optimal behavior of the pedestrian. For example, the optimal policy for the pedestrian following the unsafe objective is to always go forward. If a pedestrian following this objective were to stutter, PBVI would assign a very low probability to that objective. In contrast, \erpbvi can better account for multiple potential paths under the same objective.


\section{Conclusion}
Model-based POMDP planners must be robust to uncertainty in planning and objective inference. In this work, we presented \erpbvi which is an entropy-regularized point-based solver for POMDPs. Our experiments on toy and automotive-inspired POMDPs show that \erpbvi is more robust to modeling errors than PBVI, and also outperforms PBVI in objective inference.

In future work, we plan to investigate the theoretical properties of the \erpbvi algorithm. Additionally, we will investigate how entropy-regularized POMDP policies can be used to solve decomposed POMDPs \cite{wray2017modia, bouton2019decomposition}. Since entropy-regularized policies are less committed to single optimal actions, they may be useful for fusing multiple utility estimates from decomposed POMDPs.





\begin{acks}
The research reported in this work was supported by Alliance Innovation Laboratory Silicon Valley, Nissan North America, Inc. This material is also based upon work supported in part by the National Science Foundation Graduate Research Fellowship under Grant No. DGE-2146755. Any opinions, findings, and conclusions or recommendations expressed in this material are those of the authors and do not necessarily reflect the views of the National Science Foundation.
\end{acks}

\printbibliography

\appendix
\section{\erpbvi Pseudocode}
\label{chapter:algorithms}
This section presents pseudocode for the \erpbvi algorithm. The algorithm shares a significant amount of code with PBVI. The high-level algorithm is shown in \cref{alg:pbvi}. The algorithm takes in a POMDP formulation $\mathcal{P}$ and regularization temperature $\lambda$. The belief point set $\mathcal{B}$ is initialized with the POMDP agent's initial belief-state. The Q-functions are initialized to a lower bound.

The algorithm iteratively improves Q-function estimates, prunes unnecessary alpha-vectors, and expands the belief set. These three key steps are repeated for a fixed number of iterations. The improvement step is shown in \cref{alg:improve} and \cref{alg:backup}. The Q-functions are backed up at each point in the belief set $\mathcal{B}$. As in PBVI, several applications of the backup step may be necessary for convergence. The backup procedure follows \cref{eq:gamma-argmax}, \cref{eq:alpha-ao-soft-update}, and \cref{eq:alpha-lookahead} to produce a new backed-up alpha vector for each Q-function. After the backup step, the Q-functions are pruned to remove alpha vectors that do not contribute to the estimate of the Q-function. This pruning procedure consists of solving a set of linear programs to identify dominated alpha vectors. For more details on alpha vector pruning, see \cite{kochenderfer2022algorithms}. Finally, the expansion step expands the belief set $\mathcal{B}$. It starts by expanding every possible next belief from each belief in $\mathcal{B}$. The farthest points from points in $\mathcal{B}$ are added to $\mathcal{B}$. An open source implementation of  \erpbvi in Julia is available\footnote[1]{\href{https://github.com/hdelecki/EntropyRegularizedPBVI.jl}{https://github.com/hdelecki/EntropyRegularizedPBVI.jl}}.

\input{algorithms/erpbvi}

\input{algorithms/improve}

\input{algorithms/backup}

\end{document}

%% file: defs.tex
\newcommand{\E}{\mathbb{E}}

\DeclareMathOperator*{\argmax}{arg\,max}

\newcommand{\mat}[1]{\vect{#1}}
\renewcommand{\vec}[1]{\vect{#1}}

\newcommand*{\defeq}{\stackrel{\text{def}}{=}}

\DeclareMathOperator*{\update}{\operatorname{Update}}
\newcommand{\erpbvi}{ERPBVI }

%% file: figures/preamble.tex
\usetikzlibrary{backgrounds}
\usetikzlibrary{arrows}
\usetikzlibrary{shapes,shapes.geometric,shapes.misc}

\tikzstyle{tikzfig}=[baseline=-0.25em,scale=0.5]

\pgfkeys{/tikz/tikzit fill/.initial=0}
\pgfkeys{/tikz/tikzit draw/.initial=0}
\pgfkeys{/tikz/tikzit shape/.initial=0}
\pgfkeys{/tikz/tikzit category/.initial=0}

\pgfdeclarelayer{edgelayer}
\pgfdeclarelayer{nodelayer}
\pgfsetlayers{background,edgelayer,nodelayer,main}

\tikzstyle{none}=[inner sep=0mm]

\newcommand{\tikzfig}[1]{%
{\tikzstyle{every picture}=[tikzfig]
\IfFileExists{#1.tikz}
  {\input{#1.tikz}}
  {%
    \IfFileExists{./figures/#1.tikz}
      {\input{./figures/#1.tikz}}
      {\tikz[baseline=-0.5em]{\node[draw=red,font=\color{red},fill=red!10!white] {\textit{#1}};}}%
  }}%
}

\newcommand{\ctikzfig}[1]{%
\begin{center}\rm
  \input{#1.tikz}
\end{center}}

\tikzstyle{circ}=[fill=white, draw=none, shape=circle, minimum size=1cm, inner sep=0pt, outer sep=0pt]

\tikzstyle{arr}=[->]

\tikzstyle{every loop}=[]

\pgfplotsset{compat=newest}
\pgfplotsset{every axis legend/.append style={%
cells={anchor=west}}
}
\usetikzlibrary{arrows}
\tikzset{>=stealth'}

\definecolor{pastelRed}{RGB}{245, 97, 92}
\definecolor{pastelBlue}{RGB}{0, 114, 178}
\definecolor{pastelGreen}{RGB}{0, 158, 115}
\definecolor{pastelPurple}{RGB}{135, 112, 254}\usepgfplotslibrary{groupplots}

\usetikzlibrary{calc,angles,positioning,intersections,quotes,decorations.markings}

\usepgfplotslibrary{colormaps}
\pgfplotsset{colormap/viridis}

\pgfmathdeclarefunction{lg10}{1}{%
    \pgfmathparse{ln(#1)/ln(10)}%
}

%% file: figures/backup.tex
\begin{tikzpicture}
	\begin{pgfonlayer}{nodelayer}
		\node [] (pbvi) at (0, 1) {$\textbf{\Large PBVI}$};

		\node [] (erpbvi) at (7, 1) {$\textbf{\Large Entropy-Regularized PBVI}$};
		\node [style=circ] (14) at (0, 0) {$U(b)$};
		\node [style=circ] (15) at (0, -2) {$\text{max}(\cdot)$};
		\node [style=circ] (16) at (-1, -4) {$Q(b', a_1)$};
		\node [style=circ] (18) at (1, -4) {$Q(b', a_{n})$};
		\node [] (20) at (0, -4) {$\cdots$};

            \node [style=circ] (19) at (5, 0) {$Q_{1}(b)$};
            \node [style=circ] (20) at (5, -2) {$\text{LSE}(\cdot)$};
            \node [style=circ] (21) at (9, 0) {$Q_{n}(b)$};
            \node [style=circ] (22) at (9, -2) {$\text{LSE}(\cdot)$};
            \node [style=circ] (23) at (4, -4) {$Q_{1}(b')$};
            \node [style=circ] (24) at (6, -4) {$Q_{n}(b')$};
            \node [style=circ] (25) at (8, -4) {$Q_{1}(b')$};
            \node [style=circ] (26) at (10, -4) {$Q_{n}(b')$};

            \node [] (dots) at (7, 0) {$\cdots$};

            \node [] (dots2) at (9, -4) {$\cdots$};
            \node [] (dots3) at (5, -4) {$\cdots$};
	\end{pgfonlayer}
	\begin{pgfonlayer}{edgelayer}
		\draw [style=arr] (16) to (15);
		\draw [style=arr] (18) to (15);
		\draw [style=arr] (15) to (14);

            \draw [style=arr] (26) to (22);
            \draw [style=arr] (25) to (22);

            \draw [style=arr] (23) to (20);
            \draw [style=arr] (24) to (20);

            \draw [style=arr] (20) to (19);
            \draw [style=arr] (22) to (21);

	\end{pgfonlayer}
\end{tikzpicture}

%% file: figures/pbvistatedist.tex
\begin{tikzpicture}[]
\begin{axis}[
  enlargelimits = false,
  axis on top,
  colormap={mycolormap}{ rgb(0cm)=(0.9598724461306475,0.9923183147134332,0.9981068934505289) rgb(1cm)=(0.9563106765726231,0.9910985559124045,0.9973853038608934) rgb(2cm)=(0.9527328903602096,0.9898690520680634,0.996666220438234) rgb(3cm)=(0.9491390659072884,0.9886297218358925,0.9959496481086513) rgb(4cm)=(0.9455291826627364,0.9873804832608938,0.9952355904775686) rgb(5cm)=(0.9419032211367155,0.9861212537755221,0.9945240497956144) rgb(6cm)=(0.938261162927531,0.984851950197687,0.993815026924136) rgb(7cm)=(0.9346029907490568,0.9835724887288201,0.9931085213003051) rgb(8cm)=(0.9309286884587586,0.982282784952016,0.9924045309018413) rgb(9cm)=(0.9272382410863136,0.9809827538302512,0.9917030522113434) rgb(10cm)=(0.9235316348628526,0.9796723097046772,0.991004080180245) rgb(11cm)=(0.9198088572508324,0.9783513662929947,0.9903076081923851) rgb(12cm)=(0.9160698969745619,0.9770198366879136,0.9896136280271873) rgb(13cm)=(0.9123147440513837,0.9756776333557003,0.9889221298224861) rgb(14cm)=(0.9085433898235447,0.9743246681348131,0.98823310203696) rgb(15cm)=(0.9047558269907628,0.9729608522346321,0.9875465314122019) rgb(16cm)=(0.9009520496435086,0.9715860962342905,0.9868624029344263) rgb(17cm)=(0.8971320532970133,0.9702003100815962,0.9861806997958055) rgb(18cm)=(0.8932958349260478,0.9688034030920726,0.9855014033554498) rgb(19cm)=(0.8894433930004482,0.9673952839480949,0.9848244931000286) rgb(20cm)=(0.8855747275214475,0.9659758606981458,0.9841499466040392) rgb(21cm)=(0.8816898400588179,0.9645450407561859,0.983477739489727) rgb(22cm)=(0.8777887337888365,0.9631027309011418,0.982807845386655) rgb(23cm)=(0.8738714135331107,0.9616488372765238,0.9821402358909329) rgb(24cm)=(0.869937885798274,0.9601832653901686,0.9814748805241101) rgb(25cm)=(0.8659881588165884,0.9587059201141191,0.9808117466917264) rgb(26cm)=(0.8620222425874599,0.9572167056846391,0.9801507996415392) rgb(27cm)=(0.858040148919902,0.955715525702375,0.9794920024214163) rgb(28cm)=(0.854041891475971,0.954202283132661,0.9788353158369062) rgb(29cm)=(0.8500274858152057,0.9526768803059852,0.9781806984084903) rgb(30cm)=(0.8459969494400758,0.9511392189186064,0.9775281063285209) rgb(31cm)=(0.8419503018424922,0.9495892000333387,0.9768774934178331) rgb(32cm)=(0.8378875645514021,0.9480267240805104,0.9762288110820633) rgb(33cm)=(0.8338087611814742,0.946451690859088,0.975582008267656) rgb(34cm)=(0.8297139174829443,0.9448639995379936,0.9749370314175637) rgb(35cm)=(0.8256030613926192,0.9432635486576042,0.974293824426654) rgb(36cm)=(0.8214762230860864,0.9416502361314429,0.9736523285968207) rgb(37cm)=(0.817333435031163,0.9400239592480782,0.9730124825917975) rgb(38cm)=(0.8131747320426111,0.9383846146732259,0.972374222391689) rgb(39cm)=(0.8090001513381664,0.9367320984520692,0.9717374812472143) rgb(40cm)=(0.8048097325958974,0.9350663060117957,0.9711021896336595) rgb(41cm)=(0.8006035180129655,0.9333871321643721,0.9704682752045611) rgb(42cm)=(0.7963815523657847,0.9316944711095487,0.9698356627451025) rgb(43cm)=(0.792143883071658,0.9299882164381136,0.9692042741252395) rgb(44cm)=(0.7878905602519085,0.928268261135401,0.968574028252542) rgb(45cm)=(0.7836216367965588,0.9265344975850573,0.9679448410247774) rgb(46cm)=(0.7793371684306061,0.9247868175730848,0.9673166252822106) rgb(47cm)=(0.7750372137819259,0.9230251122921601,0.966689290759637) rgb(48cm)=(0.7707218344508712,0.9212492723462442,0.9660627440381525) rgb(49cm)=(0.7663910950815942,0.9194591877554915,0.965436888496641) rgb(50cm)=(0.7620450634351679,0.9176547479614698,0.9648116242630066) rgb(51cm)=(0.7576838104645369,0.9158358418326973,0.9641868481651287) rgb(52cm)=(0.7533074103913615,0.9140023576705104,0.9635624536815474) rgb(53cm)=(0.748915940784821,0.9121541832152773,0.9629383308918805) rgb(54cm)=(0.7445094826424122,0.9102912056529556,0.9623143664269665) rgb(55cm)=(0.7400881204728308,0.908413311622025,0.9616904434187311) rgb(56cm)=(0.7356519423809713,0.9065203872207885,0.9610664414497739) rgb(57cm)=(0.731201040155132,0.9046123180150684,0.9604422365026796) rgb(58cm)=(0.7267355093564831,0.902688989046303,0.9598177009090422) rgb(59cm)=(0.7222554494108581,0.9007502848400585,0.9591927032981927) rgb(60cm)=(0.717760963702957,0.898796089414972,0.9585671085456443) rgb(61cm)=(0.7132521596730321,0.8968262862921409,0.957940777721228) rgb(62cm)=(0.7087291489161072,0.8948407585049672,0.9573135680369239) rgb(63cm)=(0.7041920472838555,0.8928393886094819,0.9566853327943784) rgb(64cm)=(0.6996409749891771,0.8908220586951563,0.9560559213320995) rgb(65cm)=(0.6950760567135761,0.8887886503962223,0.9554251789723213) rgb(66cm)=(0.6904974217174298,0.886739044903517,0.9547929469675213) rgb(67cm)=(0.6859052039532371,0.8846731229768716,0.9541590624465991) rgb(68cm)=(0.6812995421819241,0.8825907649580557,0.9535233583606767) rgb(69cm)=(0.6766805800923308,0.8804918507843058,0.9528856634285311) rgb(70cm)=(0.6720484664239583,0.8783762600024524,0.9522458020816357) rgb(71cm)=(0.6674033550930851,0.8762438717836621,0.951603594408796) rgb(72cm)=(0.6627454053223509,0.8740945649388282,0.9509588561003649) rgb(73cm)=(0.6580747817739442,0.8719282179346126,0.9503113983920165) rgb(74cm)=(0.653391654686468,0.8697447089101836,0.9496610280080721) rgb(75cm)=(0.6486962000156303,0.8675439156946537,0.9490075471043392) rgb(76cm)=(0.6439885995788667,0.8653257158252529,0.9483507532104601) rgb(77cm)=(0.6392690412040212,0.863089986566258,0.9476904391717357) rgb(78cm)=(0.6345377188822178,0.8608366049287075,0.94702639309041) rgb(79cm)=(0.6297948329250389,0.8585654476909206,0.9463583982663791) rgb(80cm)=(0.6250405901261706,0.8562763914198624,0.9456862331373065) rgb(81cm)=(0.6202752039276304,0.8539693124933662,0.9450096712181094) rgb(82cm)=(0.6154988945907344,0.8516440871232596,0.944328481039789) rgb(83cm)=(0.610711889371947,0.8493005913794142,0.9436424260875649) rgb(84cm)=(0.6059144227037608,0.8469387012147552,0.9429512647382972) rgb(85cm)=(0.601106736380765,0.8445582924912644,0.9422547501971291) rgb(86cm)=(0.5962890797510634,0.8421592410070069,0.941552630433342) rgb(87cm)=(0.5914617099131898,0.8397414225242246,0.9408446481153611) rgb(88cm)=(0.5866248919187098,0.837304712798527,0.9401305405448751) rgb(89cm)=(0.5817788989806507,0.8348489876092183,0.939410039590027) rgb(90cm)=(0.5769240126879532,0.8323741227908057,0.938682871617619) rgb(91cm)=(0.5720605232261038,0.8298799942657218,0.9379487574242864) rgb(92cm)=(0.567188729604139,0.827366478078313,0.9372074121665843) rgb(93cm)=(0.5623089398881874,0.8248334504301286,0.9364585452899292) rgb(94cm)=(0.5574214714417353,0.8222807877165669,0.9357018604563359) rgb(95cm)=(0.5525266511728079,0.8197083665649142,0.9349370554708822) rgb(96cm)=(0.5476248157882279,0.817116063873837,0.9341638222068399) rgb(97cm)=(0.5427163120551604,0.814503756854372,0.9333818465293995) rgb(98cm)=(0.5378014970700963,0.8118713230724676,0.9325908082179072) rgb(99cm)=(0.5328807385354902,0.8092186404931362,0.9317903808865448) rgb(100cm)=(0.5279544150441919,0.8065455875262674,0.9309802319033641) rgb(101cm)=(0.5230229163718746,0.803852043074166,0.9301600223075916) rgb(102cm)=(0.5180866437776148,0.8011378865808789,0.9293294067251053) rgb(103cm)=(0.5131460103127881,0.798402998083366,0.9284880332819923) rgb(104cm)=(0.5082014411384356,0.7956472582645897,0.9276355435160879) rgb(105cm)=(0.5032533738512363,0.7928705485085842,0.926771572286376) rgb(106cm)=(0.4983022588182254,0.7900727509575846,0.9258957476801506) rgb(107cm)=(0.4933485595203659,0.7872537485712798,0.9250076909178181) rgb(108cm)=(0.48839275290507117,0.7844134251882697,0.9241070162551993) rgb(109cm)=(0.48343532974775943,0.7815516655898112,0.9231933308832246) rgb(110cm)=(0.47847679502248014,0.7786683555659247,0.9222662348248574) rgb(111cm)=(0.4735176682816455,0.7757633819839538,0.9213253208291144) rgb(112cm)=(0.4685584840448553,0.7728366328596675,0.9203701742620326) rgb(113cm)=(0.46359979219675557,0.7698879974309855,0.9194003729943933) rgb(114cm)=(0.45864215839386263,0.7669173662344408,0.9184154872860731) rgb(115cm)=(0.45368616448018545,0.7639246311844559,0.9174150796667988) rgb(116cm)=(0.4487324089114761,0.7609096856555513,0.916398704813153) rgb(117cm)=(0.44378150718783377,0.7578724245675869,0.9153659094215977) rgb(118cm)=(0.43883409229433556,0.7548127444741382,0.9143162320773257) rgb(119cm)=(0.4338908151492963,0.751730543654137,0.9132492031187178) rgb(120cm)=(0.42895234505964125,0.7486257222068752,0.9121643444971546) rgb(121cm)=(0.4240193701828045,0.7454981821505078,0.9110611696319434) rgb(122cm)=(0.41909259799443976,0.7423478275241729,0.9099391832601126) rgb(123cm)=(0.4141727557611033,0.7391745644938653,0.9087978812807714) rgb(124cm)=(0.40926059101692785,0.7359783014621867,0.907636750593772) rgb(125cm)=(0.40435687204317566,0.7327589491821298,0.9064552689323445) rgb(126cm)=(0.39946238834933884,0.7295164208750193,0.9052529046894007) rgb(127cm)=(0.3945779511543214,0.7262506323527778,0.904029116737148) rgb(128cm)=(0.38970439386597466,0.7229615021446534,0.9027833542396644) rgb(129cm)=(0.384842572557084,0.7196489516285821,0.9015150564580601) rgb(130cm)=(0.3799933664355778,0.7163129051673355,0.9002236525477989) rgb(131cm)=(0.37515767830653374,0.712953290249632,0.8989085613477892) rgb(132cm)=(0.37033643502318986,0.7095700376363773,0.8975691911607662) rgb(133cm)=(0.3655305879238609,0.7061630815122169,0.8962049395245154) rgb(134cm)=(0.36074111325130126,0.7027323596425856,0.8948151929734189) rgb(135cm)=(0.35596901255066643,0.6992778135364333,0.8933993267898125) rgb(136cm)=(0.3512153130417842,0.6957993886148311,0.8919567047445829) rgb(137cm)=(0.34648106796100836,0.6922970343856507,0.8904866788264217) rgb(138cm)=(0.34176735686744264,0.688770704624516,0.8889885889591052) rgb(139cm)=(0.3370752859077748,0.6852203575622395,0.8874617627061497) rgb(140cm)=(0.33240598803343224,0.6816459560789506,0.8859055149621363) rgb(141cm)=(0.3277606231631623,0.6780474679051318,0.884319147629971) rgb(142cm)=(0.323140378283542,0.674424865829782,0.8827019492832917) rgb(143cm)=(0.3185464674792341,0.6707781279159213,0.8810531948132129) rgb(144cm)=(0.31398013188419993,0.6671072377236673,0.8793721450585074) rgb(145cm)=(0.30944263954429974,0.663412184541106,0.8776580464183287) rgb(146cm)=(0.3049352851811019,0.6596929636231755,0.8759101304464704) rgb(147cm)=(0.3004593898459371,0.6559495764387995,0.8741276134261363) rgb(148cm)=(0.2960163004525628,0.6521820309264769,0.8723096959241304) rgb(149cm)=(0.2916073891761336,0.6483903417585689,0.8704555623232839) rgb(150cm)=(0.287234052705504,0.6445745306144802,0.8685643803319141) rgb(151cm)=(0.2828977113352822,0.6407346264629581,0.8666353004689916) rgb(152cm)=(0.27859980788359656,0.636870665853704,0.8646674555236445) rgb(153cm)=(0.2743418064210467,0.632982693218497,0.8626599599875463) rgb(154cm)=(0.2701251907960719,0.6290707611820013,0.8606119094586164) rgb(155cm)=(0.2659514629418374,0.625134930882425,0.8585223800144223) rgb(156cm)=(0.2618221409498273,0.6211752723021782,0.8563904275535293) rgb(157cm)=(0.25773875689563264,0.6171918646086464,0.8542150871029616) rgb(158cm)=(0.25370285440305884,0.6131847965051781,0.8519953720898366) rgb(159cm)=(0.24971598593364747,0.6091541665923458,0.8497302735750965) rgb(160cm)=(0.24577970978997393,0.6051000837395115,0.8474187594471637) rgb(161cm)=(0.24189558682294515,0.6010226674666752,0.8450597735732058) rgb(162cm)=(0.23806517683555684,0.5969220483365351,0.8426522349055544) rgb(163cm)=(0.2342900346783952,0.5927983683566352,0.8401950365406878) rgb(164cm)=(0.23057170603561994,0.5886517813914106,0.8376870447280442) rgb(165cm)=(0.22691172290423278,0.5844824535838474,0.8351270978257307) rgb(166cm)=(0.2233115987742032,0.5802905637864096,0.8325140052000849) rgb(167cm)=(0.21977282352250094,0.5760763040007663,0.8298465460658102) rgb(168cm)=(0.21629685804030258,0.5718398798257444,0.8271234682632483) rgb(169cm)=(0.2128851286195761,0.5675815109128078,0.8243434869691546) rgb(170cm)=(0.20953902113288436,0.563301431428205,0.821505283337104) rgb(171cm)=(0.20625987504854254,0.5589998905207697,0.8186075030634735) rgb(172cm)=(0.2030489773321424,0.5546771527941707,0.8156487548746852) rgb(173cm)=(0.19990755629476298,0.5503334987821841,0.8126276089311555) rgb(174cm)=(0.1968367754578813,0.5459692254253425,0.8095425951431666) rgb(175cm)=(0.19383772751479106,0.5415846465470283,0.8063922013935566) rgb(176cm)=(0.19091142847805875,0.5371800933267808,0.8031748716618968) rgb(177cm)=(0.18805881211200798,0.5327559147682659,0.7998890040444983) rgb(178cm)=(0.18528072475809024,0.5283124781589548,0.7965329486642865) rgb(179cm)=(0.18257789744858524,0.5238501684960555,0.7931050229182722) rgb(180cm)=(0.1799499358680572,0.5193693422158686,0.789604272441279) rgb(181cm)=(0.1773949398257464,0.5148702981079069,0.7860307913236826) rgb(182cm)=(0.17491082542060477,0.51035332882476,0.7823847494183211) rgb(183cm)=(0.17249543206733972,0.5058187253331548,0.7786663154985768) rgb(184cm)=(0.17014652665322333,0.5012667770304629,0.774875657483364) rgb(185cm)=(0.16786180815286292,0.49669777185630953,0.7710129426616312) rgb(186cm)=(0.16563891266229333,0.492111996399574,0.7670783379167071) rgb(187cm)=(0.16347541880880445,0.48750973600106323,0.7630720099508197) rgb(188cm)=(0.16136885348881924,0.48289127485212996,0.7589941255101152) rgb(189cm)=(0.15931669788285757,0.4782568960894836,0.7548448516105289) rgb(190cm)=(0.15731639369414338,0.4736068818864488,0.7506243557648463) rgb(191cm)=(0.15536534955595424,0.46894151354090713,0.7463328062113126) rgb(192cm)=(0.15346094755223078,0.4642610715601529,0.7419703721441665) rgb(193cm)=(0.1516005497963212,0.4595658357428795,0.737537223946457) rgb(194cm)=(0.1497815050138668,0.45485608525852467,0.7330335334255542) rgb(195cm)=(0.14800115507793907,0.4501320987241754,0.7284594740517412) rgb(196cm)=(0.14625684144718779,0.4453941542792445,0.723815221200322) rgb(197cm)=(0.1445459114611245,0.44064252965811845,0.7191009523976667) rgb(198cm)=(0.14286572445057227,0.43587750226098076,0.7143168475716655) rgb(199cm)=(0.14121365762551571,0.4310993492230002,0.70946308930706) rgb(200cm)=(0.13958711170722887,0.42630834748209123,0.704539863106168) rgb(201cm)=(0.13798351627624175,0.421504773845426,0.699547357655506) rgb(202cm)=(0.136400334812559,0.41668890505490686,0.6944857650988966) rgb(203cm)=(0.13483506940931492,0.41186101785178336,0.6893552813176353) rgb(204cm)=(0.1332852651457307,0.4070213890406142,0.6841561062183429) rgb(205cm)=(0.1317485141097045,0.40217029555276773,0.6788884440291775) rgb(206cm)=(0.1302224590645459,0.3973080145096689,0.6735525036051172) rgb(207cm)=(0.12870479675833912,0.3924348232859778,0.6681484987430568) rgb(208cm)=(0.12719328087779996,0.3875509995729258,0.6626766485075596) rgb(209cm)=(0.1256857246518218,0.3826568214420032,0.6571371775681097) rgb(210cm)=(0.1241800031124434,0.37775256740922275,0.6515303165488243) rgb(211cm)=(0.122674055023472,0.3728385165001721,0.645856302391632) rgb(212cm)=(0.12116588448876794,0.3679149483160876,0.6401153787340144) rgb(213cm)=(0.11965356225378973,0.36298214310117705,0.6343077963025068) rgb(214cm)=(0.118135226715089,0.35804038181143366,0.628433813323242) rgb(215cm)=(0.11660908465318501,0.35308994618518813,0.6224936959509563) rgb(216cm)=(0.11507341170468657,0.34813111881565456,0.6164877187179901) rgb(217cm)=(0.11352655258956307,0.34316418322573516,0.6104161650049753) rgb(218cm)=(0.11196692110931919,0.3381894239453567,0.6042793275350559) rgb(219cm)=(0.11039299993125362,0.3332071265916227,0.5980775088936865) rgb(220cm)=(0.10880334017329163,0.32821757795207257,0.59181102207626) rgb(221cm)=(0.1071965608028827,0.32322106607135337,0.5854801910660499) rgb(222cm)=(0.10557134786225547,0.31821788034161297,0.5790853514452489) rgb(223cm)=(0.10392645353086397,0.3132083115969393,0.5726268510421704) rgb(224cm)=(0.10226069503431551,0.308192652212177,0.566105050618069) rgb(225cm)=(0.10057295340714792,0.3031711962064542,0.5595203245974263) rgb(226cm)=(0.09886217211483614,0.2981442393517747,0.5528730618460389) rgb(227cm)=(0.09712735553806834,0.2931120792870168,0.5461636665017923) rgb(228cm)=(0.09536756731980514,0.28807501563769555,0.5393925588636495) rgb(229cm)=(0.09358192857274247,0.28303335014183517,0.5325601763451152) rgb(230cm)=(0.09176961594154953,0.27798738678229945,0.5256669744993266) rgb(231cm)=(0.08992985951057542,0.2729374319259088,0.5187134281239263) rgb(232cm)=(0.08806194054352859,0.26788379446965055,0.511700032455076) rgb(233cm)=(0.08616518903663528,0.2628267859942725,0.5046273044614051) rgb(234cm)=(0.0842389810611682,0.25776672092547837,0.4974957842503412) rgb(235cm)=(0.08228273586438292,0.2527039167029111,0.49030603660130584) rgb(236cm)=(0.08029591268988465,0.2476386939570015,0.48305865264263265) rgb(237cm)=(0.07827800726864173,0.24257137669366302,0.4757542516919577) rgb(238cm)=(0.07622854792014541,0.23750229248665128,0.4683934832832984) rgb(239cm)=(0.07414709118864637,0.23243177267722148,0.4609770294082654) rgb(240cm)=(0.07203321692152859,0.227360152580447,0.4535056070039693) rgb(241cm)=(0.06988652267460338,0.22228777169723896,0.4459799707264956) rgb(242cm)=(0.06770661730107405,0.21721497393065808,0.43840091605655906) rgb(243cm)=(0.06549311354550816,0.2121421078045473,0.43076928279354126) rgb(244cm)=(0.06324561941901705,0.20706952668176798,0.4230859590060839) rgb(245cm)=(0.06096372807366051,0.20199758897836334,0.4153518855223839) rgb(246cm)=(0.05864700581896331,0.19692665836869766,0.40756806106223736) rgb(247cm)=(0.05629497782486128,0.19185710397497655,0.3997355481368923) rgb(248cm)=(0.05390711092568983,0.18678930053236933,0.39185547987345665) rgb(249cm)=(0.05148279276692084,0.18172362851807133,0.38392906796017434) rgb(250cm)=(0.04902130630376956,0.17666047422882786,0.3759576119602476) rgb(251cm)=(0.04652179834471041,0.17160022978631376,0.36794251030914754) rgb(252cm)=(0.04398324039784387,0.16654329304288917,0.35988527339921467) rgb(253cm)=(0.04140437947230644,0.16149006735090626,0.3517875392738329) rgb(254cm)=(0.03877591103302971,0.15644096114598163,0.3436510926130204) },
  colorbar,
  xmin = 0.5,
  xmax = 5.5,
  ymin = 0.5,
  ymax = 5.5,
  height = 8cm,
  width =8cm,
  ticks=none,
  xtick={0.5, 1.5, 2.5, 3.5, 4.5},,
  ytick={0.5, 1.5, 2.5, 3.5, 4.5},
  grid=both,
  grid style={line width=.5pt, draw=black},
  major grid style={line width=0.5pt,draw=black}
]

\addplot[
  point meta min = 0.0,
  point meta max = 1.00,
  point meta = explicit,
  matrix plot*,
  mesh/cols = 5,
  mesh/rows = 5
] table[
  meta = data
] {figures/pbvi_states.dat};

\end{axis}
\end{tikzpicture}

%% file: figures/erstatedist.tex
\begin{tikzpicture}[]
\begin{axis}[
  enlargelimits = false,
  axis on top,
  colormap={mycolormap}{ rgb(0cm)=(0.9598724461306475,0.9923183147134332,0.9981068934505289) rgb(1cm)=(0.9563106765726231,0.9910985559124045,0.9973853038608934) rgb(2cm)=(0.9527328903602096,0.9898690520680634,0.996666220438234) rgb(3cm)=(0.9491390659072884,0.9886297218358925,0.9959496481086513) rgb(4cm)=(0.9455291826627364,0.9873804832608938,0.9952355904775686) rgb(5cm)=(0.9419032211367155,0.9861212537755221,0.9945240497956144) rgb(6cm)=(0.938261162927531,0.984851950197687,0.993815026924136) rgb(7cm)=(0.9346029907490568,0.9835724887288201,0.9931085213003051) rgb(8cm)=(0.9309286884587586,0.982282784952016,0.9924045309018413) rgb(9cm)=(0.9272382410863136,0.9809827538302512,0.9917030522113434) rgb(10cm)=(0.9235316348628526,0.9796723097046772,0.991004080180245) rgb(11cm)=(0.9198088572508324,0.9783513662929947,0.9903076081923851) rgb(12cm)=(0.9160698969745619,0.9770198366879136,0.9896136280271873) rgb(13cm)=(0.9123147440513837,0.9756776333557003,0.9889221298224861) rgb(14cm)=(0.9085433898235447,0.9743246681348131,0.98823310203696) rgb(15cm)=(0.9047558269907628,0.9729608522346321,0.9875465314122019) rgb(16cm)=(0.9009520496435086,0.9715860962342905,0.9868624029344263) rgb(17cm)=(0.8971320532970133,0.9702003100815962,0.9861806997958055) rgb(18cm)=(0.8932958349260478,0.9688034030920726,0.9855014033554498) rgb(19cm)=(0.8894433930004482,0.9673952839480949,0.9848244931000286) rgb(20cm)=(0.8855747275214475,0.9659758606981458,0.9841499466040392) rgb(21cm)=(0.8816898400588179,0.9645450407561859,0.983477739489727) rgb(22cm)=(0.8777887337888365,0.9631027309011418,0.982807845386655) rgb(23cm)=(0.8738714135331107,0.9616488372765238,0.9821402358909329) rgb(24cm)=(0.869937885798274,0.9601832653901686,0.9814748805241101) rgb(25cm)=(0.8659881588165884,0.9587059201141191,0.9808117466917264) rgb(26cm)=(0.8620222425874599,0.9572167056846391,0.9801507996415392) rgb(27cm)=(0.858040148919902,0.955715525702375,0.9794920024214163) rgb(28cm)=(0.854041891475971,0.954202283132661,0.9788353158369062) rgb(29cm)=(0.8500274858152057,0.9526768803059852,0.9781806984084903) rgb(30cm)=(0.8459969494400758,0.9511392189186064,0.9775281063285209) rgb(31cm)=(0.8419503018424922,0.9495892000333387,0.9768774934178331) rgb(32cm)=(0.8378875645514021,0.9480267240805104,0.9762288110820633) rgb(33cm)=(0.8338087611814742,0.946451690859088,0.975582008267656) rgb(34cm)=(0.8297139174829443,0.9448639995379936,0.9749370314175637) rgb(35cm)=(0.8256030613926192,0.9432635486576042,0.974293824426654) rgb(36cm)=(0.8214762230860864,0.9416502361314429,0.9736523285968207) rgb(37cm)=(0.817333435031163,0.9400239592480782,0.9730124825917975) rgb(38cm)=(0.8131747320426111,0.9383846146732259,0.972374222391689) rgb(39cm)=(0.8090001513381664,0.9367320984520692,0.9717374812472143) rgb(40cm)=(0.8048097325958974,0.9350663060117957,0.9711021896336595) rgb(41cm)=(0.8006035180129655,0.9333871321643721,0.9704682752045611) rgb(42cm)=(0.7963815523657847,0.9316944711095487,0.9698356627451025) rgb(43cm)=(0.792143883071658,0.9299882164381136,0.9692042741252395) rgb(44cm)=(0.7878905602519085,0.928268261135401,0.968574028252542) rgb(45cm)=(0.7836216367965588,0.9265344975850573,0.9679448410247774) rgb(46cm)=(0.7793371684306061,0.9247868175730848,0.9673166252822106) rgb(47cm)=(0.7750372137819259,0.9230251122921601,0.966689290759637) rgb(48cm)=(0.7707218344508712,0.9212492723462442,0.9660627440381525) rgb(49cm)=(0.7663910950815942,0.9194591877554915,0.965436888496641) rgb(50cm)=(0.7620450634351679,0.9176547479614698,0.9648116242630066) rgb(51cm)=(0.7576838104645369,0.9158358418326973,0.9641868481651287) rgb(52cm)=(0.7533074103913615,0.9140023576705104,0.9635624536815474) rgb(53cm)=(0.748915940784821,0.9121541832152773,0.9629383308918805) rgb(54cm)=(0.7445094826424122,0.9102912056529556,0.9623143664269665) rgb(55cm)=(0.7400881204728308,0.908413311622025,0.9616904434187311) rgb(56cm)=(0.7356519423809713,0.9065203872207885,0.9610664414497739) rgb(57cm)=(0.731201040155132,0.9046123180150684,0.9604422365026796) rgb(58cm)=(0.7267355093564831,0.902688989046303,0.9598177009090422) rgb(59cm)=(0.7222554494108581,0.9007502848400585,0.9591927032981927) rgb(60cm)=(0.717760963702957,0.898796089414972,0.9585671085456443) rgb(61cm)=(0.7132521596730321,0.8968262862921409,0.957940777721228) rgb(62cm)=(0.7087291489161072,0.8948407585049672,0.9573135680369239) rgb(63cm)=(0.7041920472838555,0.8928393886094819,0.9566853327943784) rgb(64cm)=(0.6996409749891771,0.8908220586951563,0.9560559213320995) rgb(65cm)=(0.6950760567135761,0.8887886503962223,0.9554251789723213) rgb(66cm)=(0.6904974217174298,0.886739044903517,0.9547929469675213) rgb(67cm)=(0.6859052039532371,0.8846731229768716,0.9541590624465991) rgb(68cm)=(0.6812995421819241,0.8825907649580557,0.9535233583606767) rgb(69cm)=(0.6766805800923308,0.8804918507843058,0.9528856634285311) rgb(70cm)=(0.6720484664239583,0.8783762600024524,0.9522458020816357) rgb(71cm)=(0.6674033550930851,0.8762438717836621,0.951603594408796) rgb(72cm)=(0.6627454053223509,0.8740945649388282,0.9509588561003649) rgb(73cm)=(0.6580747817739442,0.8719282179346126,0.9503113983920165) rgb(74cm)=(0.653391654686468,0.8697447089101836,0.9496610280080721) rgb(75cm)=(0.6486962000156303,0.8675439156946537,0.9490075471043392) rgb(76cm)=(0.6439885995788667,0.8653257158252529,0.9483507532104601) rgb(77cm)=(0.6392690412040212,0.863089986566258,0.9476904391717357) rgb(78cm)=(0.6345377188822178,0.8608366049287075,0.94702639309041) rgb(79cm)=(0.6297948329250389,0.8585654476909206,0.9463583982663791) rgb(80cm)=(0.6250405901261706,0.8562763914198624,0.9456862331373065) rgb(81cm)=(0.6202752039276304,0.8539693124933662,0.9450096712181094) rgb(82cm)=(0.6154988945907344,0.8516440871232596,0.944328481039789) rgb(83cm)=(0.610711889371947,0.8493005913794142,0.9436424260875649) rgb(84cm)=(0.6059144227037608,0.8469387012147552,0.9429512647382972) rgb(85cm)=(0.601106736380765,0.8445582924912644,0.9422547501971291) rgb(86cm)=(0.5962890797510634,0.8421592410070069,0.941552630433342) rgb(87cm)=(0.5914617099131898,0.8397414225242246,0.9408446481153611) rgb(88cm)=(0.5866248919187098,0.837304712798527,0.9401305405448751) rgb(89cm)=(0.5817788989806507,0.8348489876092183,0.939410039590027) rgb(90cm)=(0.5769240126879532,0.8323741227908057,0.938682871617619) rgb(91cm)=(0.5720605232261038,0.8298799942657218,0.9379487574242864) rgb(92cm)=(0.567188729604139,0.827366478078313,0.9372074121665843) rgb(93cm)=(0.5623089398881874,0.8248334504301286,0.9364585452899292) rgb(94cm)=(0.5574214714417353,0.8222807877165669,0.9357018604563359) rgb(95cm)=(0.5525266511728079,0.8197083665649142,0.9349370554708822) rgb(96cm)=(0.5476248157882279,0.817116063873837,0.9341638222068399) rgb(97cm)=(0.5427163120551604,0.814503756854372,0.9333818465293995) rgb(98cm)=(0.5378014970700963,0.8118713230724676,0.9325908082179072) rgb(99cm)=(0.5328807385354902,0.8092186404931362,0.9317903808865448) rgb(100cm)=(0.5279544150441919,0.8065455875262674,0.9309802319033641) rgb(101cm)=(0.5230229163718746,0.803852043074166,0.9301600223075916) rgb(102cm)=(0.5180866437776148,0.8011378865808789,0.9293294067251053) rgb(103cm)=(0.5131460103127881,0.798402998083366,0.9284880332819923) rgb(104cm)=(0.5082014411384356,0.7956472582645897,0.9276355435160879) rgb(105cm)=(0.5032533738512363,0.7928705485085842,0.926771572286376) rgb(106cm)=(0.4983022588182254,0.7900727509575846,0.9258957476801506) rgb(107cm)=(0.4933485595203659,0.7872537485712798,0.9250076909178181) rgb(108cm)=(0.48839275290507117,0.7844134251882697,0.9241070162551993) rgb(109cm)=(0.48343532974775943,0.7815516655898112,0.9231933308832246) rgb(110cm)=(0.47847679502248014,0.7786683555659247,0.9222662348248574) rgb(111cm)=(0.4735176682816455,0.7757633819839538,0.9213253208291144) rgb(112cm)=(0.4685584840448553,0.7728366328596675,0.9203701742620326) rgb(113cm)=(0.46359979219675557,0.7698879974309855,0.9194003729943933) rgb(114cm)=(0.45864215839386263,0.7669173662344408,0.9184154872860731) rgb(115cm)=(0.45368616448018545,0.7639246311844559,0.9174150796667988) rgb(116cm)=(0.4487324089114761,0.7609096856555513,0.916398704813153) rgb(117cm)=(0.44378150718783377,0.7578724245675869,0.9153659094215977) rgb(118cm)=(0.43883409229433556,0.7548127444741382,0.9143162320773257) rgb(119cm)=(0.4338908151492963,0.751730543654137,0.9132492031187178) rgb(120cm)=(0.42895234505964125,0.7486257222068752,0.9121643444971546) rgb(121cm)=(0.4240193701828045,0.7454981821505078,0.9110611696319434) rgb(122cm)=(0.41909259799443976,0.7423478275241729,0.9099391832601126) rgb(123cm)=(0.4141727557611033,0.7391745644938653,0.9087978812807714) rgb(124cm)=(0.40926059101692785,0.7359783014621867,0.907636750593772) rgb(125cm)=(0.40435687204317566,0.7327589491821298,0.9064552689323445) rgb(126cm)=(0.39946238834933884,0.7295164208750193,0.9052529046894007) rgb(127cm)=(0.3945779511543214,0.7262506323527778,0.904029116737148) rgb(128cm)=(0.38970439386597466,0.7229615021446534,0.9027833542396644) rgb(129cm)=(0.384842572557084,0.7196489516285821,0.9015150564580601) rgb(130cm)=(0.3799933664355778,0.7163129051673355,0.9002236525477989) rgb(131cm)=(0.37515767830653374,0.712953290249632,0.8989085613477892) rgb(132cm)=(0.37033643502318986,0.7095700376363773,0.8975691911607662) rgb(133cm)=(0.3655305879238609,0.7061630815122169,0.8962049395245154) rgb(134cm)=(0.36074111325130126,0.7027323596425856,0.8948151929734189) rgb(135cm)=(0.35596901255066643,0.6992778135364333,0.8933993267898125) rgb(136cm)=(0.3512153130417842,0.6957993886148311,0.8919567047445829) rgb(137cm)=(0.34648106796100836,0.6922970343856507,0.8904866788264217) rgb(138cm)=(0.34176735686744264,0.688770704624516,0.8889885889591052) rgb(139cm)=(0.3370752859077748,0.6852203575622395,0.8874617627061497) rgb(140cm)=(0.33240598803343224,0.6816459560789506,0.8859055149621363) rgb(141cm)=(0.3277606231631623,0.6780474679051318,0.884319147629971) rgb(142cm)=(0.323140378283542,0.674424865829782,0.8827019492832917) rgb(143cm)=(0.3185464674792341,0.6707781279159213,0.8810531948132129) rgb(144cm)=(0.31398013188419993,0.6671072377236673,0.8793721450585074) rgb(145cm)=(0.30944263954429974,0.663412184541106,0.8776580464183287) rgb(146cm)=(0.3049352851811019,0.6596929636231755,0.8759101304464704) rgb(147cm)=(0.3004593898459371,0.6559495764387995,0.8741276134261363) rgb(148cm)=(0.2960163004525628,0.6521820309264769,0.8723096959241304) rgb(149cm)=(0.2916073891761336,0.6483903417585689,0.8704555623232839) rgb(150cm)=(0.287234052705504,0.6445745306144802,0.8685643803319141) rgb(151cm)=(0.2828977113352822,0.6407346264629581,0.8666353004689916) rgb(152cm)=(0.27859980788359656,0.636870665853704,0.8646674555236445) rgb(153cm)=(0.2743418064210467,0.632982693218497,0.8626599599875463) rgb(154cm)=(0.2701251907960719,0.6290707611820013,0.8606119094586164) rgb(155cm)=(0.2659514629418374,0.625134930882425,0.8585223800144223) rgb(156cm)=(0.2618221409498273,0.6211752723021782,0.8563904275535293) rgb(157cm)=(0.25773875689563264,0.6171918646086464,0.8542150871029616) rgb(158cm)=(0.25370285440305884,0.6131847965051781,0.8519953720898366) rgb(159cm)=(0.24971598593364747,0.6091541665923458,0.8497302735750965) rgb(160cm)=(0.24577970978997393,0.6051000837395115,0.8474187594471637) rgb(161cm)=(0.24189558682294515,0.6010226674666752,0.8450597735732058) rgb(162cm)=(0.23806517683555684,0.5969220483365351,0.8426522349055544) rgb(163cm)=(0.2342900346783952,0.5927983683566352,0.8401950365406878) rgb(164cm)=(0.23057170603561994,0.5886517813914106,0.8376870447280442) rgb(165cm)=(0.22691172290423278,0.5844824535838474,0.8351270978257307) rgb(166cm)=(0.2233115987742032,0.5802905637864096,0.8325140052000849) rgb(167cm)=(0.21977282352250094,0.5760763040007663,0.8298465460658102) rgb(168cm)=(0.21629685804030258,0.5718398798257444,0.8271234682632483) rgb(169cm)=(0.2128851286195761,0.5675815109128078,0.8243434869691546) rgb(170cm)=(0.20953902113288436,0.563301431428205,0.821505283337104) rgb(171cm)=(0.20625987504854254,0.5589998905207697,0.8186075030634735) rgb(172cm)=(0.2030489773321424,0.5546771527941707,0.8156487548746852) rgb(173cm)=(0.19990755629476298,0.5503334987821841,0.8126276089311555) rgb(174cm)=(0.1968367754578813,0.5459692254253425,0.8095425951431666) rgb(175cm)=(0.19383772751479106,0.5415846465470283,0.8063922013935566) rgb(176cm)=(0.19091142847805875,0.5371800933267808,0.8031748716618968) rgb(177cm)=(0.18805881211200798,0.5327559147682659,0.7998890040444983) rgb(178cm)=(0.18528072475809024,0.5283124781589548,0.7965329486642865) rgb(179cm)=(0.18257789744858524,0.5238501684960555,0.7931050229182722) rgb(180cm)=(0.1799499358680572,0.5193693422158686,0.789604272441279) rgb(181cm)=(0.1773949398257464,0.5148702981079069,0.7860307913236826) rgb(182cm)=(0.17491082542060477,0.51035332882476,0.7823847494183211) rgb(183cm)=(0.17249543206733972,0.5058187253331548,0.7786663154985768) rgb(184cm)=(0.17014652665322333,0.5012667770304629,0.774875657483364) rgb(185cm)=(0.16786180815286292,0.49669777185630953,0.7710129426616312) rgb(186cm)=(0.16563891266229333,0.492111996399574,0.7670783379167071) rgb(187cm)=(0.16347541880880445,0.48750973600106323,0.7630720099508197) rgb(188cm)=(0.16136885348881924,0.48289127485212996,0.7589941255101152) rgb(189cm)=(0.15931669788285757,0.4782568960894836,0.7548448516105289) rgb(190cm)=(0.15731639369414338,0.4736068818864488,0.7506243557648463) rgb(191cm)=(0.15536534955595424,0.46894151354090713,0.7463328062113126) rgb(192cm)=(0.15346094755223078,0.4642610715601529,0.7419703721441665) rgb(193cm)=(0.1516005497963212,0.4595658357428795,0.737537223946457) rgb(194cm)=(0.1497815050138668,0.45485608525852467,0.7330335334255542) rgb(195cm)=(0.14800115507793907,0.4501320987241754,0.7284594740517412) rgb(196cm)=(0.14625684144718779,0.4453941542792445,0.723815221200322) rgb(197cm)=(0.1445459114611245,0.44064252965811845,0.7191009523976667) rgb(198cm)=(0.14286572445057227,0.43587750226098076,0.7143168475716655) rgb(199cm)=(0.14121365762551571,0.4310993492230002,0.70946308930706) rgb(200cm)=(0.13958711170722887,0.42630834748209123,0.704539863106168) rgb(201cm)=(0.13798351627624175,0.421504773845426,0.699547357655506) rgb(202cm)=(0.136400334812559,0.41668890505490686,0.6944857650988966) rgb(203cm)=(0.13483506940931492,0.41186101785178336,0.6893552813176353) rgb(204cm)=(0.1332852651457307,0.4070213890406142,0.6841561062183429) rgb(205cm)=(0.1317485141097045,0.40217029555276773,0.6788884440291775) rgb(206cm)=(0.1302224590645459,0.3973080145096689,0.6735525036051172) rgb(207cm)=(0.12870479675833912,0.3924348232859778,0.6681484987430568) rgb(208cm)=(0.12719328087779996,0.3875509995729258,0.6626766485075596) rgb(209cm)=(0.1256857246518218,0.3826568214420032,0.6571371775681097) rgb(210cm)=(0.1241800031124434,0.37775256740922275,0.6515303165488243) rgb(211cm)=(0.122674055023472,0.3728385165001721,0.645856302391632) rgb(212cm)=(0.12116588448876794,0.3679149483160876,0.6401153787340144) rgb(213cm)=(0.11965356225378973,0.36298214310117705,0.6343077963025068) rgb(214cm)=(0.118135226715089,0.35804038181143366,0.628433813323242) rgb(215cm)=(0.11660908465318501,0.35308994618518813,0.6224936959509563) rgb(216cm)=(0.11507341170468657,0.34813111881565456,0.6164877187179901) rgb(217cm)=(0.11352655258956307,0.34316418322573516,0.6104161650049753) rgb(218cm)=(0.11196692110931919,0.3381894239453567,0.6042793275350559) rgb(219cm)=(0.11039299993125362,0.3332071265916227,0.5980775088936865) rgb(220cm)=(0.10880334017329163,0.32821757795207257,0.59181102207626) rgb(221cm)=(0.1071965608028827,0.32322106607135337,0.5854801910660499) rgb(222cm)=(0.10557134786225547,0.31821788034161297,0.5790853514452489) rgb(223cm)=(0.10392645353086397,0.3132083115969393,0.5726268510421704) rgb(224cm)=(0.10226069503431551,0.308192652212177,0.566105050618069) rgb(225cm)=(0.10057295340714792,0.3031711962064542,0.5595203245974263) rgb(226cm)=(0.09886217211483614,0.2981442393517747,0.5528730618460389) rgb(227cm)=(0.09712735553806834,0.2931120792870168,0.5461636665017923) rgb(228cm)=(0.09536756731980514,0.28807501563769555,0.5393925588636495) rgb(229cm)=(0.09358192857274247,0.28303335014183517,0.5325601763451152) rgb(230cm)=(0.09176961594154953,0.27798738678229945,0.5256669744993266) rgb(231cm)=(0.08992985951057542,0.2729374319259088,0.5187134281239263) rgb(232cm)=(0.08806194054352859,0.26788379446965055,0.511700032455076) rgb(233cm)=(0.08616518903663528,0.2628267859942725,0.5046273044614051) rgb(234cm)=(0.0842389810611682,0.25776672092547837,0.4974957842503412) rgb(235cm)=(0.08228273586438292,0.2527039167029111,0.49030603660130584) rgb(236cm)=(0.08029591268988465,0.2476386939570015,0.48305865264263265) rgb(237cm)=(0.07827800726864173,0.24257137669366302,0.4757542516919577) rgb(238cm)=(0.07622854792014541,0.23750229248665128,0.4683934832832984) rgb(239cm)=(0.07414709118864637,0.23243177267722148,0.4609770294082654) rgb(240cm)=(0.07203321692152859,0.227360152580447,0.4535056070039693) rgb(241cm)=(0.06988652267460338,0.22228777169723896,0.4459799707264956) rgb(242cm)=(0.06770661730107405,0.21721497393065808,0.43840091605655906) rgb(243cm)=(0.06549311354550816,0.2121421078045473,0.43076928279354126) rgb(244cm)=(0.06324561941901705,0.20706952668176798,0.4230859590060839) rgb(245cm)=(0.06096372807366051,0.20199758897836334,0.4153518855223839) rgb(246cm)=(0.05864700581896331,0.19692665836869766,0.40756806106223736) rgb(247cm)=(0.05629497782486128,0.19185710397497655,0.3997355481368923) rgb(248cm)=(0.05390711092568983,0.18678930053236933,0.39185547987345665) rgb(249cm)=(0.05148279276692084,0.18172362851807133,0.38392906796017434) rgb(250cm)=(0.04902130630376956,0.17666047422882786,0.3759576119602476) rgb(251cm)=(0.04652179834471041,0.17160022978631376,0.36794251030914754) rgb(252cm)=(0.04398324039784387,0.16654329304288917,0.35988527339921467) rgb(253cm)=(0.04140437947230644,0.16149006735090626,0.3517875392738329) rgb(254cm)=(0.03877591103302971,0.15644096114598163,0.3436510926130204) },
  colorbar,
  xmin = 0.5,
  xmax = 5.5,
  ymin = 0.5,
  ymax = 5.5,
  height = 8cm,
  width =8cm,
  ticks=none,
  xtick={0.5, 1.5, 2.5, 3.5, 4.5},,
  ytick={0.5, 1.5, 2.5, 3.5, 4.5},
  grid=both,
  grid style={line width=.5pt, draw=black},
  major grid style={line width=0.5pt,draw=black}
]

\addplot[
  point meta min = 0.0,
  point meta max = 1.0,
  point meta = explicit,
  matrix plot*,
  mesh/cols = 5,
  mesh/rows = 5
] table[
  meta = data
] {figures/erpbvi_states.dat};

\end{axis}
\end{tikzpicture}

%% file: figures/tiger_robustness_single.tex
\begin{tikzpicture}[]
\begin{axis}[
  legend pos = {south west},
  ylabel = {$\mathbb{E}\left[ R\right]$},
  xlabel = {$\lambda$},
  xmode = {log}
]


\addplot+[
  mark = {none},
  solid, pastelBlue, opacity=1.0, mark options={fill=pastelBlue, opacity=1.0}, thick
] coordinates {
  (0.010000000000000002, 32.9253337067875)
  (0.01373823795883263, 32.831528647908364)
  (0.018873918221350976, 32.96546908220783)
  (0.02592943797404667, 32.64085142998186)
  (0.035622478902624426, 32.717378925150854)
  (0.04893900918477494, 32.7908081028375)
  (0.06723357536499339, 33.0726747987165)
  (0.09236708571873861, 33.060876955455086)
  (0.12689610031679222, 33.04648098583554)
  (0.17433288221999882, 32.716197448076066)
  (0.23950266199874853, 32.42604602968059)
  (0.3290344562312668, 31.955145315009933)
  (0.4520353656360243, 30.478382205585692)
  (0.6210169418915615, 29.82384585169669)
  (0.8531678524172809, 29.205785482377973)
  (1.1721022975334803, 28.42709101850564)
  (1.6102620275609394, 27.932698243774077)
  (2.2122162910704493, 26.50105778449044)
  (3.0391953823131974, 24.733076832450113)
  (4.1753189365604015, 21.69626975228556)
  (5.736152510448679, -373.81400070591985)
  (7.880462815669913, -475.5008132052081)
  (10.826367338740546, -526.4311714748873)
  (14.87352107293511, -554.7114689139909)
  (20.433597178569418, -570.5822366115287)
  (28.072162039411772, -581.7359836806568)
  (38.56620421163472, -589.9825095456353)
  (52.98316906283707, -593.8114849144108)
  (72.7895384398315, -598.54287580359)
  (100.0, -597.8789926387967)
};
\addlegendentry{{}{$p_{correct}=0.90$}}


\addplot+[
  mark = {none},
  solid, black, opacity=1.0, mark options={fill=black, opacity=1.0}, thick
] coordinates {
  (0.010000000000000002, 19.12625277238971)
  (0.01373823795883263, 19.187021718898613)
  (0.018873918221350976, 19.706922406729298)
  (0.02592943797404667, 18.519519029705023)
  (0.035622478902624426, 19.059555057757525)
  (0.04893900918477494, 19.569375551793648)
  (0.06723357536499339, 19.589794630832063)
  (0.09236708571873861, 19.4396337048676)
  (0.12689610031679222, 19.003929344436447)
  (0.17433288221999882, 19.543642988260217)
  (0.23950266199874853, 18.803431118116798)
  (0.3290344562312668, 18.894730559168856)
  (0.4520353656360243, 19.16460848130835)
  (0.6210169418915615, 18.54105581509553)
  (0.8531678524172809, 17.86077794526504)
  (1.1721022975334803, 17.987350315863885)
  (1.6102620275609394, 17.21915354888824)
  (2.2122162910704493, 16.06329599580511)
  (3.0391953823131974, 14.320846663769323)
  (4.1753189365604015, 9.519159286681289)
  (5.736152510448679, -383.7226587741583)
  (7.880462815669913, -484.44178704437587)
  (10.826367338740546, -531.57419625559)
  (14.87352107293511, -556.1400848449247)
  (20.433597178569418, -574.2687469226772)
  (28.072162039411772, -581.737395689382)
  (38.56620421163472, -587.3581394975051)
  (52.98316906283707, -593.9589371812399)
  (72.7895384398315, -596.4852142467483)
  (100.0, -598.249707375754)
};
\addlegendentry{{}{$p_{correct}=0.85$}}

\addplot+[
  mark = {none},
  solid, pastelGreen, opacity=1.0, mark options={fill=pastelGreen, opacity=1.0}, thick
] coordinates {
  (0.010000000000000002, -45.222150538789606)
  (0.01373823795883263, -45.07028208647669)
  (0.018873918221350976, -45.83396188593598)
  (0.02592943797404667, -45.973234867020516)
  (0.035622478902624426, -45.9130103748809)
  (0.04893900918477494, -45.148641656928)
  (0.06723357536499339, -45.639628801490716)
  (0.09236708571873861, -44.81154490691432)
  (0.12689610031679222, -45.24927305022229)
  (0.17433288221999882, -45.88814480429533)
  (0.23950266199874853, -43.55278493820891)
  (0.3290344562312668, -42.10854259829541)
  (0.4520353656360243, -39.527236582288424)
  (0.6210169418915615, -37.095503849244274)
  (0.8531678524172809, -35.729171535017706)
  (1.1721022975334803, -34.22640230376476)
  (1.6102620275609394, -33.82628753784475)
  (2.2122162910704493, -35.135415095221354)
  (3.0391953823131974, -38.63567631149103)
  (4.1753189365604015, -46.28388743100809)
  (5.736152510448679, -408.9807238374968)
  (7.880462815669913, -499.6568389133625)
  (10.826367338740546, -541.8104887430347)
  (14.87352107293511, -565.6227128868715)
  (20.433597178569418, -577.5408063817167)
  (28.072162039411772, -586.2665677270438)
  (38.56620421163472, -592.0744126699419)
  (52.98316906283707, -595.1975238186557)
  (72.7895384398315, -596.7985410714628)
  (100.0, -599.0706379917476)
};
\addlegendentry{{}{$p_{correct}=0.70$}}


\addplot+[
  mark = {none},
  solid, pastelRed, opacity=1.0, mark options={fill=pastelRed, opacity=1.0}, thick
] coordinates {
  (0.010000000000000002, -108.75522282389663)
  (0.01373823795883263, -107.07840790557363)
  (0.018873918221350976, -107.70544834951396)
  (0.02592943797404667, -107.8868829456493)
  (0.035622478902624426, -107.74325981420502)
  (0.04893900918477494, -107.94591576937819)
  (0.06723357536499339, -107.76815371577307)
  (0.09236708571873861, -107.44712247877652)
  (0.12689610031679222, -106.29704232283211)
  (0.17433288221999882, -106.74474025908562)
  (0.23950266199874853, -105.86308951500364)
  (0.3290344562312668, -103.05865262500991)
  (0.4520353656360243, -97.46676606861134)
  (0.6210169418915615, -93.40206784458208)
  (0.8531678524172809, -89.577505502929)
  (1.1721022975334803, -88.0950911172469)
  (1.6102620275609394, -86.53169041944271)
  (2.2122162910704493, -85.86738689804696)
  (3.0391953823131974, -90.90672853033269)
  (4.1753189365604015, -100.12820830529684)
  (5.736152510448679, -431.27995157356986)
  (7.880462815669913, -514.7056222737225)
  (10.826367338740546, -552.2506803914691)
  (14.87352107293511, -566.765238322032)
  (20.433597178569418, -579.34380324932)
  (28.072162039411772, -591.7571359378388)
  (38.56620421163472, -596.5032996193912)
  (52.98316906283707, -596.4249818345875)
  (72.7895384398315, -598.701449596109)
  (100.0, -602.5960821489126)
};
\addlegendentry{{}{$p_{correct}=0.60$}}

\addplot+[
  dashed, no marks, pastelRed, opacity=1.0
] coordinates {
  (0.010000000000000002, -108.49461093845221)
  (100.0, -108.49461093845221)
};

\addplot+[
  dashed, no marks, pastelGreen, opacity=1.0
] coordinates {
  (0.010000000000000002, -45.65420670627665)
  (100.0, -45.65420670627665)
};

\addplot+[
  dashed, no marks, black, opacity=1.0
] coordinates {
  (0.010000000000000002, 19.27210843036081)
  (100.0, 19.27210843036081)
};

\addplot+[
  dashed, no marks, pastelBlue, opacity=1.0
] coordinates {
  (0.010000000000000002, 33.06930534082997)
  (100.0, 33.06930534082997)
};

\end{axis}

\end{tikzpicture}

%% file: figures/grid_pareto.tex
\begin{tikzpicture}[]
\begin{axis}[
  legend pos = {north east},
  ylabel = {$\mathbb{E}\left[ R\right]$},
  xlabel = {$\lambda$},
  xmode = {log}
]

\addplot+[
  mark = {none},
  solid, black, opacity=1.0, mark options={fill=black, opacity=1.0}, thick
] coordinates {
  (0.0001, 0.8735454973079895)
  (0.0001610262027560939, 0.8735454973079895)
  (0.0002592943797404667, 0.8735454973079895)
  (0.00041753189365604, 0.8735454973079895)
  (0.0006723357536499335, 0.8735454973079895)
  (0.001082636733874054, 0.8735454973079895)
  (0.0017433288221999873, 0.8735439984715916)
  (0.0028072162039411755, 0.873325187092961)
  (0.004520353656360245, 0.870787385669981)
  (0.007278953843983154, 0.8609392209892761)
  (0.0117210229753348, 0.8306980328046873)
  (0.018873918221350976, 0.7626347273635422)
  (0.03039195382313198, 0.6460649079408952)
  (0.04893900918477494, 0.4736384109624178)
  (0.07880462815669913, 0.2681641956082662)
  (0.12689610031679222, 0.06376217978953636)
  (0.20433597178569418, -0.0013497755130776738)
  (0.3290344562312668, 0.008827803465652822)
  (0.5298316906283709, -0.16058810870227996)
  (0.8531678524172809, -0.7659641490826927)
  (1.3738237958832629, -0.9484244388838747)
  (2.2122162910704493, -0.9789997898107355)
  (3.562247890262442, -0.9919827772932516)
  (5.736152510448679, -0.9978279920284534)
  (9.236708571873862, -0.9998712716297511)
  (14.87352107293511, -1.0026341984482474)
  (23.95026619987486, -1.000293671635285)
  (38.56620421163472, -0.9999158204182989)
  (62.10169418915616, -0.9997629640560284)
  (100.0, -1.0005347532141975)
};
\addlegendentry{{}{$p_{slip}=0.0$}}

\addplot+[
  mark = {none},
  solid, pastelBlue, opacity=1.0, mark options={fill=pastelBlue, opacity=1.0}, thick
] coordinates {
  (0.0001, 0.09902183404250568)
  (0.0001610262027560939, 0.09927173378128538)
  (0.0002592943797404667, 0.10092187129163065)
  (0.00041753189365604, 0.10276957078753647)
  (0.0006723357536499335, 0.10483070763521962)
  (0.001082636733874054, 0.11004894979761644)
  (0.0017433288221999873, 0.09913925651484932)
  (0.0028072162039411755, 0.10982604415655155)
  (0.004520353656360245, 0.11566694936225479)
  (0.007278953843983154, 0.12041978284876906)
  (0.0117210229753348, 0.13309328788804098)
  (0.018873918221350976, 0.06894868781726347)
  (0.03039195382313198, -0.08742744589277192)
  (0.04893900918477494, -0.2805924167386706)
  (0.07880462815669913, -0.4537141752948542)
  (0.12689610031679222, -0.5783804325439874)
  (0.20433597178569418, -0.6169951785277509)
  (0.3290344562312668, -0.6017209908335589)
  (0.5298316906283709, -0.5895799511965518)
  (0.8531678524172809, -0.8576040654434048)
  (1.3738237958832629, -0.9631848471869467)
  (2.2122162910704493, -0.9859279043283663)
  (3.562247890262442, -0.9943642812561324)
  (5.736152510448679, -0.9980930796748723)
  (9.236708571873862, -0.9991426939525507)
  (14.87352107293511, -1.0010907056669918)
  (23.95026619987486, -0.9989963271134804)
  (38.56620421163472, -0.9997996825513524)
  (62.10169418915616, -0.9999586415589998)
  (100.0, -1.0001532857284365)
};
\addlegendentry{{}{$p_{slip}=0.1$}}

\addplot+[
  mark = {none},
  solid, pastelGreen, opacity=1.0, mark options={fill=pastelGreen, opacity=1.0}, thick
] coordinates {
  (0.0001, -0.6890495329260404)
  (0.0001610262027560939, -0.6915306348317789)
  (0.0002592943797404667, -0.6941122521648101)
  (0.00041753189365604, -0.6955027797985205)
  (0.0006723357536499335, -0.6953689140319942)
  (0.001082636733874054, -0.6819684641968008)
  (0.0017433288221999873, -0.6787757984611821)
  (0.0028072162039411755, -0.6599952249549629)
  (0.004520353656360245, -0.6590536455252797)
  (0.007278953843983154, -0.6394479288920365)
  (0.0117210229753348, -0.6035953535983679)
  (0.018873918221350976, -0.6047432250205409)
  (0.03039195382313198, -0.6663088216931262)
  (0.04893900918477494, -0.7520619603151627)
  (0.07880462815669913, -0.8184873832181847)
  (0.12689610031679222, -0.8627838527666442)
  (0.20433597178569418, -0.8759364227205404)
  (0.3290344562312668, -0.8675496886651493)
  (0.5298316906283709, -0.8440546629964648)
  (0.8531678524172809, -0.9293835093839679)
  (1.3738237958832629, -0.9783199049442491)
  (2.2122162910704493, -0.9899429356436472)
  (3.562247890262442, -0.9943129017499348)
  (5.736152510448679, -0.9970666816321891)
  (9.236708571873862, -0.9984342580454574)
  (14.87352107293511, -1.0008725498898978)
  (23.95026619987486, -0.9982118414641142)
  (38.56620421163472, -0.9987635774867019)
  (62.10169418915616, -0.9986323202085627)
  (100.0, -0.9983973722590135)
};
\addlegendentry{{}{$p_{slip}=0.3$}}

\addplot+[
  mark = {none},
  solid, pastelRed, opacity=1.0, mark options={fill=pastelRed, opacity=1.0}, thick
] coordinates {
  (0.0001, -0.9369129090126034)
  (0.0001610262027560939, -0.940874381114548)
  (0.0002592943797404667, -0.9398692955677618)
  (0.00041753189365604, -0.9371666964806988)
  (0.0006723357536499335, -0.94073205411445)
  (0.001082636733874054, -0.9389089735701155)
  (0.0017433288221999873, -0.9340898180477621)
  (0.0028072162039411755, -0.9277215814363261)
  (0.004520353656360245, -0.925398134955251)
  (0.007278953843983154, -0.9148976183106232)
  (0.0117210229753348, -0.8941966309502785)
  (0.018873918221350976, -0.8792245806871333)
  (0.03039195382313198, -0.8838433361036148)
  (0.04893900918477494, -0.9033816294694801)
  (0.07880462815669913, -0.9238644046670565)
  (0.12689610031679222, -0.934394083280651)
  (0.20433597178569418, -0.9383346119413586)
  (0.3290344562312668, -0.9340176615683493)
  (0.5298316906283709, -0.9316420440876633)
  (0.8531678524172809, -0.9626880867233115)
  (1.3738237958832629, -0.9846128547088736)
  (2.2122162910704493, -0.9897567875285705)
  (3.562247890262442, -0.9928410299233724)
  (5.736152510448679, -0.9952803189766446)
  (9.236708571873862, -0.9970088740829713)
  (14.87352107293511, -0.9980011685275646)
  (23.95026619987486, -0.9931667169268894)
  (38.56620421163472, -0.9926133848868718)
  (62.10169418915616, -0.9936207643929128)
  (100.0, -0.9921752659208505)
};
\addlegendentry{{}{$p_{slip}=0.5$}}

\addplot+[
  dashed, no marks, black, opacity=1.0
] coordinates {
  (0.0001, 0.8735454973079895)
  (100.0, 0.8735454973079895)
};

\addplot+[
  dashed, no marks, pastelBlue, opacity=1.0
] coordinates {
  (0.0001, 0.1022738967440203)
  (100.0, 0.1022738967440203)
};

\addplot+[
  dashed, no marks, pastelGreen, opacity=1.0
] coordinates {
  (0.0001, -0.6916071306710612)
  (100.0, -0.6916071306710612)
};

\addplot+[
  dashed, no marks, pastelRed, opacity=1.0
] coordinates {
  (0.0001, -0.9382469867947751)
  (100.0, -0.9382469867947751)
};

\end{axis}
\end{tikzpicture}

%% file: figures/gridworld_inference_tpr_fpr.tex
\begin{tikzpicture}[]
\begin{axis}[
  colorbar = {true},
  legend pos = {south east},
  ylabel = {True Positive Rate},
  xmin = {0},
  xmax = {1},
  ymax = {1},
  xlabel = {False Positive Rate},
  ymin = {0},
  colorbar style={yticklabel=$10^{\pgfmathparse{\tick}\pgfmathprintnumber\pgfmathresult}$, ylabel=$\lambda$ (temperature)}
]

\addplot+[
  draw = none,
  mark = {*},
  mark options={fill=black},
  only marks, black
] coordinates {
  (2, 2)
  (2, 2)
};
\addlegendentry{{}{PBVI}}

\addplot+[
  draw = none,
  mark = {square*},
  mark options={fill=black},
  only marks, black, fill=black
] coordinates {
  (2, 2)
  (2, 2)
};
\addlegendentry{{}{ERPBVI}}

\addplot+[
  scatter,
  scatter src = explicit,
  only marks = {true},
  mark = {*},
] coordinates {
  (0.43400000000000005, 0.548) [-2.0]
  (0.42600000000000005, 0.569) [-1.7894736842105263]
  (0.41700000000000004, 0.545) [-1.5789473684210527]
  (0.44299999999999995, 0.592) [-1.368421052631579]
  (0.44199999999999995, 0.559) [-1.1578947368421053]
  (0.44099999999999995, 0.561) [-0.9473684210526315]
  (0.43999999999999995, 0.544) [-0.7368421052631579]
  (0.42800000000000005, 0.572) [-0.5263157894736842]
  (0.44699999999999995, 0.584) [-0.3157894736842105]
  (0.44599999999999995, 0.546) [-0.10526315789473684]
  (0.47, 0.563) [0.10526315789473681]
  (0.47, 0.565) [0.3157894736842105]
  (0.507, 0.55) [0.5263157894736842]
  (0.509, 0.564) [0.7368421052631579]
  (0.491, 0.572) [0.9473684210526315]
  (0.515, 0.513) [1.1578947368421053]
  (0.478, 0.52) [1.368421052631579]
  (0.517, 0.505) [1.5789473684210527]
  (0.52, 0.504) [1.7894736842105263]
  (0.491, 0.474) [2.0]
};

\addplot+[
  scatter,
  scatter src = explicit,
  only marks = {true},
  mark={square*}
] coordinates {
  (0.43799999999999994, 0.835) [-2.0]
  (0.28900000000000003, 0.791) [-1.7894736842105263]
  (0.239, 0.641) [-1.5789473684210527]
  (0.259, 0.656) [-1.368421052631579]
  (0.346, 0.699) [-1.1578947368421053]
  (0.33999999999999997, 0.664) [-0.9473684210526315]
  (0.33299999999999996, 0.658) [-0.7368421052631579]
  (0.31200000000000006, 0.631) [-0.5263157894736842]
  (0.344, 0.655) [-0.3157894736842105]
  (0.345, 0.677) [-0.10526315789473684]
  (0.371, 0.688) [0.10526315789473681]
  (0.362, 0.681) [0.3157894736842105]
  (0.363, 0.693) [0.5263157894736842]
  (0.43799999999999994, 0.689) [0.7368421052631579]
  (0.45999999999999996, 0.711) [0.9473684210526315]
  (0.394, 0.651) [1.1578947368421053]
  (0.43300000000000005, 0.593) [1.368421052631579]
  (0.469, 0.521) [1.5789473684210527]
  (0.46399999999999997, 0.49) [1.7894736842105263]
  (0.504, 0.502) [2.0]
};

\addplot+[
  mark = {none},
  dashed, black, 
] coordinates {
  (0, 0)
  (1, 1)
};

\end{axis}
\end{tikzpicture}

%% file: figures/crosswalk_tpr.tex
\begin{tikzpicture}[]
\begin{axis}[
  colorbar = {true},
  legend pos = {south east},
  ylabel = {True Positive Rate},
  xmin = {0},
  xmax = {1},
  ymax = {1},
  xlabel = {False Positive Rate},
  ymin = {0},
  colorbar style={yticklabel=$10^{\pgfmathparse{\tick}\pgfmathprintnumber\pgfmathresult}$, ylabel=$\lambda$ (temperature)}
]

\addplot+[
  draw = none,
  mark = {*},
  mark options={fill=black},
  only marks, black
] coordinates {
  (2, 2)
  (2, 2)
};
\addlegendentry{{}{PBVI}}

\addplot+[
  draw = none,
  mark = {square*},
  mark options={fill=black},
  only marks, black, fill=black
] coordinates {
  (2, 2)
  (2, 2)
};
\addlegendentry{{}{ERPBVI}}

\addplot+[
  scatter,
  scatter src = explicit,
  mark = {*},
  only marks = {true}
] coordinates {
  (0.44699999999999995, 0.533) [-4.0]
  (0.0, 0.029) [4.0]
  (0.15800000000000003, 0.428) [-2.0]
  (0.0, 0.029) [1.0]
  (0.0, 0.029) [-1.0]
  (0.393, 0.495) [-3.0]
  (0.0, 0.029) [3.0]
  (0.0, 0.029) [0.0]
  (0.0, 0.029) [2.0]
};

\addplot+[
  scatter,
  scatter src = explicit,
  mark = {square*},
  only marks = {true}
] coordinates {
  (0.39, 0.594) [-4.0]
  (0.504, 0.51) [4.0]
  (0.583, 0.918) [-2.0]
  (0.20399999999999996, 0.838) [1.0]
  (0.522, 0.911) [-1.0]
  (0.556, 0.904) [-3.0]
  (0.33699999999999997, 0.805) [3.0]
  (0.40700000000000003, 0.869) [0.0]
  (0.17700000000000005, 0.818) [2.0]
};


\addplot+[
  mark = {none},
  dashed, black, 
] coordinates {
  (0, 0)
  (1, 1)
};

\end{axis}
\end{tikzpicture}

%% file: algorithms/erpbvi.tex
\begin{algorithm}[H]
    \small
    \caption{Entropy-regularized Point-based Value Iteration}
    \label{alg:pbvi}
    \begin{algorithmic}[1]
        \Require $\mathcal{P} \defeq \langle \mathcal{S}, \mathcal{A}, \mathcal{O}, T, R, O, \gamma \rangle$: POMDP
        \Require $\lambda$: Temperature
        \Function{ERPBVI$(\mathcal{P}, \lambda)$}{}
            \State $B \leftarrow \textproc{InitializeBeliefs}(\mathcal{P})$
            \State $\Xi \leftarrow \textproc{InitializeQValues}(\mathcal{P})$
            \For {$i \leftarrow 1 \textbf{ to } n_\text{iterations}$}
                \State $\Xi \leftarrow {\textproc{Improve}}(\mathcal{P}, B, \Xi)$ \Comment{Q-function improvement}
                \State $\Xi \leftarrow \textproc{Prune}(\Xi)$ \Comment{Q-value pruning}
                \State $B \leftarrow \textproc{Expand}(\mathcal{P}, B)$ \Comment{Belief expansion}
            \EndFor
            \State \Return $\pi(\Xi)$ \Comment{Alpha-vector policy}
        \EndFunction
    \end{algorithmic}
\end{algorithm}

%% file: algorithms/improve.tex
\begin{algorithm}[H]
    \small
    \caption{Q-value Improvement}
    \label{alg:improve}
    \begin{algorithmic}[1]
        \Function{${\textproc{Improve}}(\mathcal{P}, B, \Xi)$}{}
            \For {$\vec{b} \text{ in } B$}
                \State $\Xi \leftarrow {\textproc{Backup}}(\mathcal{P}, \vec{b}, \Xi)$
                    
            \EndFor

            \State \Return $\Xi$
        \EndFunction
    \end{algorithmic}
\end{algorithm}

%% file: algorithms/backup.tex
\begin{algorithm}[H]
    \small
    \caption{Backup}
    \label{alg:backup}
    \begin{algorithmic}[1]
        \Function{${\textproc{Backup}}(\mathcal{P}, \vec{b}, \Xi)$}{}
            \For {$a \text{ in } \mathcal{A}$}
                \For {$o \text{ in } \mathcal{O}$}
                    \State $\vec{b}' \leftarrow {\textproc{Update}}(\vec{b}, a, o)$
                    \For{$\Gamma_i \text{ in } \Xi$}
                        \State $A_{a, o}(i) \leftarrow {\textproc{argmax}}(\Gamma_i, \vec{b}')$ \Comment{\cref{eq:gamma-argmax}}
                    \EndFor
                    \State $\alpha_{a,o} \leftarrow A_{a, o} {\textproc{SoftMax}}(A_{a,o}^{\top } \vec{b}'/\lambda)$ \Comment{\cref{eq:alpha-ao-soft-update}}
                \EndFor
                \State $\alpha_a \leftarrow {\textproc{lookahead}}(\mathcal{P}, \alpha_{a, o})$ \Comment{\cref{eq:alpha-lookahead}}
                \State ${\textproc{append}}(\Xi, \alpha_a)$
            \EndFor
            \State \Return $\Xi$
        \EndFunction
    \end{algorithmic}
\end{algorithm}